\documentclass{egpubl}
\usepackage{egsr2025_DLOnlyTrack}
\title{Content-Aware Texturing for Gaussian Splatting}
\author[Panagiotis Papantonakis \& Georgios Kopanas \& Frédo Durand \& George Drettakis]
{\parbox{\textwidth}{\centering Panagiotis Papantonakis$^{1,2}$\orcid{0009-0009-5058-1249},
        Georgios Kopanas\thanks{work done at Google}$^{3,4}$\orcid{0009-0002-5829-2192},
        Frédo Durand$^{5}$\orcid{0000-0001-9919-069X}
        and George Drettakis$^{1,2}$\orcid{0000-0002-9254-4819}
}\\
{\parbox{\textwidth}{\centering $^1$ Inria $^2$ Université Côte D'Azur $^3$ Google $^4$ Runway ML
         $^5$MIT CSAIL
       }
   }
}

\SpecialIssuePaper         
\WsPaper           

\CGFccby

\usepackage[T1]{fontenc}
\usepackage{dfadobe}  

\biberVersion
\BibtexOrBiblatex
\usepackage[backend=biber,bibstyle=EG,citestyle=alphabetic,backref=true]{biblatex} 
\electronicVersion
\PrintedOrElectronic

\ifpdf \usepackage[pdftex]{graphicx} \pdfcompresslevel=9
\else \usepackage[dvips]{graphicx} \fi

\usepackage{egweblnk} 
\usepackage{amssymb, amsmath}
\usepackage{booktabs}
\usepackage{multirow}
\usepackage[table]{xcolor}
\usepackage{tabularx}
\usepackage{graphicx}
\usepackage{tikz}
\usetikzlibrary{spy}
\usepackage{rotating}
\usepackage{soul}

\usepackage[ruled]{algorithm2e}

\begin{document}
\teaser{
 \includegraphics[width=1\linewidth]{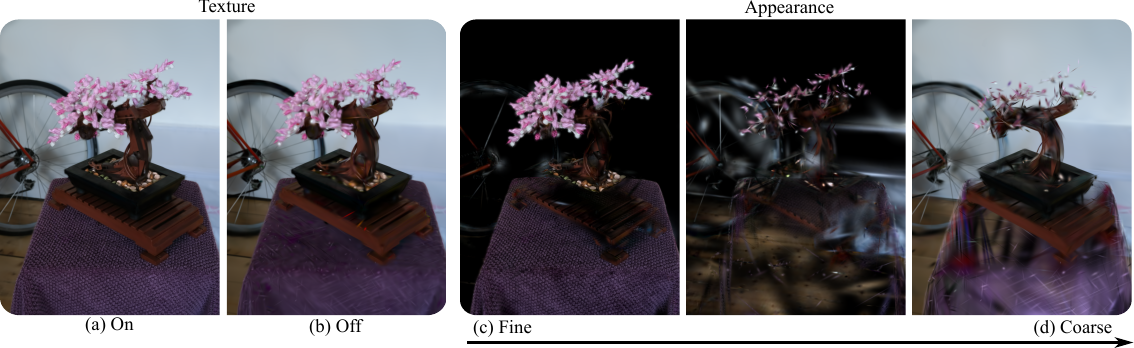}
\centering
\caption{
We propose a content-aware texturing method for 2D Gaussian Splatting.  Our textures reconstruct intricate scene detail (a). Gaussian primitives reconstruct the shape of the scene at low frequency of appearance; we show this in (b) where texturing is disabled. Our method is adaptive, allowing different primitives to have different texel sizes, depending on scene content. On the right hand panel we display primitives with progressively higher \emph{texel-to-pixel} ratio. In regions with high frequency appearance, texels have size close that of pixels (e.g., the table cover (c)). For the low-frequency walls however (d), the ratio is high, with each texel representing a large number of input image pixels.
}
\label{fig:teaser}
}

\newcommand{\missingcitation}[1]{\textcolor{magenta}{#1}}
\newcommand{\TODO}[1]{}
\newcommand{\PP}[1]{\textcolor{cyan}{#1}}
\newcommand{\GD}[1]{\textcolor{blue}{#1}}
\newcommand{\GK}[1]{\textcolor{brown}{#1}}
\newcommand{\Fredo}[1]{\textcolor{orange}{Fr\'edo: #1}}
\newcommand{\revision}[2]{#2}
\definecolor{tabfirst}{rgb}{1, 0.7, 0.7} 

\definecolor{tabsecond}{rgb}{1, 0.85, 0.7} 
\definecolor{tabthird}{rgb}{1, 1, 0.7} 

\newcommand{\mvec}[1]{\mathbf{#1}}

\newcommand{\spyimage}[5]{%
	\begin{tikzpicture}[spy using outlines={rectangle,#1,magnification=5,size=1cm, connect spies}]
		\node[anchor=south west,inner sep=0]  at (0,0) {\includegraphics[width=#2]{#3}};
		\spy on (#4) in node [left] at (#5);
	\end{tikzpicture}%
}

\maketitle
\begin{abstract}
Gaussian Splatting has become the method of choice for 3D reconstruction and real-time rendering of captured real scenes.
However, fine appearance details  need to be represented as a large number of small Gaussian primitives, which can be wasteful when geometry and appearance exhibit different frequency characteristics.
Inspired by the long tradition of texture mapping, we propose to use texture to represent detailed appearance where possible.
Our main focus is to incorporate per-primitive texture maps that adapt to the 
scene in a principled manner during Gaussian Splatting optimization.
We do this by proposing a new appearance representation for 2D Gaussian primitives with textures where the size of a texel is bounded by the image sampling frequency and adapted to the content of the input images. 
We achieve this by adaptively upscaling or downscaling the texture resolution during optimization.
In addition,
our approach enables control of the number of primitives during optimization based on texture resolution.
We show that our approach performs favorably in image quality and  total number of parameters used compared to alternative solutions for textured Gaussian primitives.
\end{abstract}

\section{Introduction}
%
%

Gaussian Splatting~\cite{3dgs,2dgs} has become the method of choice for the  capture and real-time novel-view synthesis of real scenes. It relies on flexible, easy-to-optimize \emph{Gaussian primitives} to represent the scene.
However, in the standard approach, fine visual appearance -- such as intricate texture on a surface -- needs to be represented with a large number of very small primitives even when the geometric complexity is low. 
This wasteful limitation arises because each Gaussian primitive is associated with a single color sample, preventing the decoupling of appearance and geometric complexity.
Rendering techniques, on the other hand, have traditionally built on \emph{texture mapping} to represent such detailed appearance. We build on this tradition to propose a representation for {\em textured} Gaussian primitives that is guided by the scene and its content.

In recent and concurrent work~\cite{bbsplat, gstex, textured3dgs, supergaussians} a number of methods introduce textured Gaussian primitives, 
providing a spatially varying color.
Most of these methods
use planar 2D Gaussians to match the dimensionality of traditional textures.
Each primitive carries several parameters (position, rotation, scale, opacity and spherical harmonics (SH)),
while textures are 2D arrays of RGB(A) texels.
The fundamental challenge of such approaches is how to allocate model capacity across the scene
and between the number of texels and the number of primitives used.
Previous and concurrent solutions either deal with the issue by imposing a fixed texture resolution per primitive~\cite{bbsplat, supergaussians, textured3dgs}, or propose heuristic approaches, that are far from ideal~\cite{gstex}.
This leaves open the question of how to share resources in a more principled manner.

To answer this question,
we propose a \emph{content-aware} texturing solution for 2D Gaussian primitives that uses the appearance complexity \revision{rev6315 term content unclear}{as observed from the input images of the scene} to make informed decisions on how parameter resources
are distributed.
We do this by first introducing a textured Gaussian primitive representation in which texel size is \emph{fixed} in world space.
\revision{rev6315 term content unclear}{
This makes the appearance texture of each primitive independent of its size,
and therefore unaffected by the growing or shrinking that it undergoes during optimization.
This representation separates the parameters of geometry and appearance, allowing us to refine them independently with appropriate allocation of memory capacity.
}
More specifically,
we introduce a method to increase and decrease the texel size as optimization evolves,
by analyzing the frequency content that primitive textures can represent and determining the corresponding error.
We pair that with a resolution-aware method to control the overall number of primitives,
striking a balance between texture size and number of primitives.
Our solution interacts closely with the optimization,
using the downscale/upscale mechanism to reduce the error in \emph{appearance},
while our control of the number of primitives handles \emph{geometric} error by spawning additional Gaussians to capture geometric detail.

%

\noindent
In summary,
we propose three contributions:
\begin{itemize}
\item A texture representation for 2D Gaussians that defines texels with a \emph{fixed} world space size, supporting visual reconstruction independent of the shape of the primitives.
\item A progressive algorithm that adaptively determines texel size and allows for content-aware fitting of the scene.
\item A resolution-based solution to control the number of primitives,
adjusted to the textured representation.

\end{itemize}
Our experiments show that our method provides a good balance between visual quality and the number of parameters used for the representation, comparing favorably to other textured Gaussian primitives solutions proposed in recent and concurrent work.
\section{Related Work}

\subsection{Traditional Representations}


\emph{Texture Mapping} was introduced in the early days of CG~\cite{catmull1974subdivision} and is widely used to effectively map detailed appearance information onto a simpler geometric representation, such as triangles. Textures are stored as 2D image arrays with a mapping function between the 3D surface and 2D texture coordinates. Computing this 2D surface parameterization is a well-studied and difficult problem~\cite{hormann2007mesh, li2018optcuts, srinivasan2024nuvo}. Intrinsic 2D parameterizations pose challenges even in traditional graphics pippelines~\cite{yuksel2019rethinking} because they introduce difficulties in content creation and produce visual artifacts in rendering. In the context of optimization for inverse problems -- such as novel view synthesis -- intrinsic parameterizations are also challenging since the geometry is not known in advance~\cite{srinivasan2024nuvo}. An alternative, non-parametric way to store localized information on the surface is by attaching it on the primitives themselves, i.e.,  attaching colors to the vertices of a mesh, requiring a very fine mesh subdivision to represent details. Mesh Colors~\cite{yuksel2010mesh,mallett2020patch} extend this idea by using more color samples along the edges and the faces of the triangles. This significantly improves the representational capacity of non-parametric appearance textures. While non-parametric appearance models are limited in many ways, they have significant advantages in the context of optimization since they overcome the need to jointly solve for texture parameterization and the surface. Our representation is also a non-parametric texture representation for Gaussian Splatting.

\subsection{Representations for 3D Reconstruction}
\label{sec:reps}
Scene reconstruction and novel view synthesis has recently utilized differentiable rendering to recover 3D representations from a set of images or video. Early methods utilized Multi-Plane Images (MPIs)~\cite{mildenhall2019local,zhou2018stereo}, a collection of planes with RGBA textures that are re-projected and rendered very efficiently using homography transformations. The semi-transparent and planar nature of the RGBA textures allowed the optimization to converge to useful 3D representations, but the planar assumption for the geometry is insufficient in most realistic cases. Neural Radiance Fields (NeRF~\cite{nerf}) popularized volumetric representations in the context of 3D reconstruction by introducing an Multi-Layer Perceptron (MLP) that stores volume density and color. NeRF's success resulted in follow-up work that extended it to support anti-aliasing and unbounded scenes~\cite{mipnerf, mipnerf360, zhang2020nerf++}, improve its training and rendering speed~\cite{kilonerf, dvxgo, instant-ngp, tensorf, plenoxels, zipnerf, adaptiveshells, sharma2024volumetric}, increase its robustness when using fewer views \cite{pixelnerf, regnerf}, and better reconstruct surfaces and handle reflections~\cite{refnerf, neuralangelo, bakedsdf}. Most closely related to our work is Nuvo~\cite{srinivasan2024nuvo} that recovers a 2D parameterization of a volume. For a more complete survey we refer readers to~\cite{tewari2022advances}.

Gaussian Splatting~\cite{3dgs} has emerged as an alternative, point-based approach to NeRFs by replacing the volumetric field with a collection of semi-transparent ellipsoidal primitives. This technique achieves excellent quality and real-time rendering even at high resolutions.
Several methods built on top of 3DGS to provide anti-aliasing \cite{mipsplatting},
by changing the appearance model but are constrained by the 1-to-1 relationship between color samples and primitives~\cite{specgaussian,malarz2025gaussian}. 

2D Gaussian Splatting~\cite{2dgs} flattens one of the dimensions of the ellipsoids to create planar discs or \emph{surfels} to align better with surfaces. 
Surfels are parametrized by the set of parameters $A=\{\boldsymbol{\mu}, \boldsymbol {\sigma}, \mvec{q}, o, \mvec{SH}\}$,
where $\boldsymbol{\mu} \in \mathbb{R}^3$ is the primitive's center,
$\boldsymbol{\sigma} \in \mathbb{R^+}^2$ are the scales of the primal axes of the surfel,
$\mvec{q} \in \mathbb{R}^4$ is a quaternion that represents the rotation $\mvec{R}$,
$o$ is the opacity and
$\mvec{SH}$ are the spherical harmonics coefficients used to get the view-dependent colour $\mvec{c}$.
The normal vector $\mvec{n}$ of the surfel is the third column of the rotation matrix $\mvec{R}$.

With this formulation, the intersection point between a ray $\mvec{r} = \mvec{r_0} + t\mvec{d}$ and a primitive can be computed as:
\begin{equation}\label{eq:intersection_point}
\mvec{p} = \mvec{r}_0 + t\mvec{d}, t = \frac{\mvec{n} \cdot (\mvec{\mu} - \mvec{r}_0)}{\mvec{n} \cdot \mvec{d}}
\end{equation}
We also build on 2D Gaussian Splatting for our method.

The initial primitive-based representations were not as compact as NeRFs, but a number of recent results allow competitive compression strategies~\cite{3dgszip}. Several recent methods build on traditional CG solutions, demonstrating that disentangling appearance from geometry can achieve significant improvements in terms of storage and memory costs.  Texture-GS~\cite{uv3dgs} combines deferred shading and a per-pixel UV coordinate to fetch appearance from a texture. This assumes that the objects can be represented by a sphere imposing topological constraints. Our results demonstrate that non-parametric texture mapping is a convenient and flexible way to recover texture during the optimization. 

Recent and concurrent work \cite{bbsplat, textured3dgs, supergaussians, gstex} have used textures to represent fine visual details.
\cite{gstex} and \cite{textured3dgs} use converged 2DGS and 3DGS point clouds, respectively,
as a starting point of their method, and then proceed to add and optimize texture parameters.
Several methods~\cite{bbsplat, textured3dgs, supergaussians} use a fixed texture resolution for each primitive that is a hyperparameter of the method. GSTex~\cite{gstex} distributes a texel budget over the primitives proportional to their size, resulting in different resolutions for different-sized primitives.
All these methods define their textures in the Gaussian canonical space, with the textures undergoing the same transformations as the primitive, which leads to texture stretching and shrinking.
SuperGaussians~\cite{supergaussians} also experiment with representing the texture with a small Neural Network.
In our approach,
we explore a different and more robust design that allows for texture maps to dynamically adjust to the captured content.
Along with two strategies that control the size of the texels and the number of primitives that are designed specifically for our representation,
we are able to reconstruct scenes without the need for a pretrained model.

\section{Method}

We first define a new representation for appearance of the Gaussian primitives that allows for content-aware texturing.
We then show how to adapt the texturing process to scene content,
by taking into account the different screen-space frequencies that need to be represented using textures.
This is achieved in a progressive manner that is compatible with the optimization process.
Finally,
we propose a resolution-based primitive management method that splits primitives allowing the geometry and appearance of the scene to be approximated as required.

In contrast to the original 2D Gaussian Splatting approach~\cite{2dgs} (see
Sec.~\ref{sec:reps} and Eq.~\ref{eq:intersection_point}), we express 
the intersection between a camera ray and a primitive in different coordinate systems, denoted by a superscript.
This is done,
as some coordinate systems make some operations easier.
Specifically:
\begin{itemize}
    \item $\mvec{p}^w = \mvec{p}$ is the intersection point in world space,
    \item $\mvec{p}^{w_0} = \mvec{p}^w - \boldsymbol{\mu}$ is in world space, centered on the primitive,
    \item $\mvec{p}^l = \mvec{R}^{-1}\mvec{p}^{w_0}$ is in the local, axis-aligned coordinate system of the primitive. Since the last component of this point is 0, we consider that $\mvec{p}^l \in \mathbb{R}^2$
    \item $\mvec{p}^{c} = \mvec{S}^{-1}\mvec{p}^l$, where $\mvec{S} = \text{diag}(\boldsymbol{\sigma})$ is in the local, normalised coordinate system of the primitive.
\end{itemize}
This last coordinate system can be considered as the ``canonical'' space of the Gaussian primitive.
The use of these spaces is purely to facilitate some operations.
For instance, \revision{rev9584 typo}{Eq.~\ref{eq:intersection_point}} is evaluated in the camera view space,
where $\mvec{r}_0$ is on top of the origin,
while anything that involves querying the texture map is better done in a coordinate frame local to the respective primitive.

 \revision{rev9584 typo}{The} color of a ray $\mvec{r}$ is computed by alpha blending:
\begin{align}
    \mvec{C(\mvec{r})} &=  \sum_i w_i(\mvec{p}_i)\mvec{c}_i(\mvec{d}) =  \sum_i T_i o_i G_i(\mvec{p}_i)\mvec{c}_i(\mvec{d}) \\
    T_i &= \prod_j^{i-1} (1-o_jG_j(\mvec{p}_j))
\end{align}
where $w_i(\mvec{p}_i)$ is the contribution of the Gaussian, 
$G(\mvec{x})$ is the evaluation of the Gaussian function that defines the falloff $G(\mvec{x}) = e^{-\frac{1}{2}({\mvec{x}^T\mvec{x}})}$, and $T_i$ is transmittance.


\subsection{Content-Aware Textures for Gaussian Splats}

We augment each 2D Gaussian primitive with a texture map 
$\mathcal{T}$.
The texture map adds a spatially varying \emph{offset} $\mvec{c}^\mathcal{T}$ to the original spherical harmonic color representation. 
The texture colors vary only spatially and have no directional dependency. This representation is compact and models only the diffuse properties of the surface, akin to albedo multiplied by irradiance. 
As a result, the color of each primitive is now also dependent on the intersection point between the ray emanating from the pixel and the 2D primitive:
\begin{equation}
    \mvec{c}_i = \mvec{SH}(\mvec{d}) + \text{bilerp}(\mathcal{T}_i, \mvec{u})
\end{equation}
where $\mvec{u}$ are the coordinates in UV space at which the ray intersects primitive $i$ with a direction vector $\mvec{d}$, and $\mvec{SH}$ are the spherical harmonics.
We need to carefully design the texture coordinate calculation
that transforms the intersection point into local, normalized primitive space $\mvec{p}^c$ to UV coordinates. 
As we discuss below, it is especially important to consider how this mapping changes when the primitive parameters are updated during optimization.

A simple mapping from the canonical space to UV texture space fixes the relative texture coordinates on the primitive such that the texture will stretch and deform with scaling and rotation of the primitive:
\begin{equation}
\label{eq:basic-u}
    \mvec{u} = \left(\frac{\mvec{p}^c}{2s_i} + 0.5\right)\cdot T_\text{res}
\end{equation}
where $s_i$ is the extent of the texture in units of standard deviations of the Gaussian primitive, and $T_\text{res}$ is the texture resolution. When the extent of the texture is smaller than the primitive, some type of padding needs to be applied.

\begin{figure}[!h]
\includegraphics[width=\linewidth]{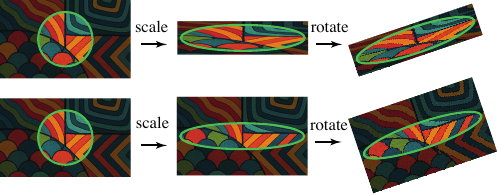}
\caption{
\label{fig:stretch}
Difference between two mappings for a primitive having learned a particular texture.
Top row: a naive approach distorts the appearance after the primitive undergoes scaling. Bottom row: in our approach, as texel size is fixed in world space, scaling the primitive only reveals more part of the underlying texture, preserving existing content.
}
\end{figure}
Recall that primitives change size and shape during the optimization process. 
As a result, with such a texture mapping, the appearance parameters of the texture are coupled with the parameters that control the shape of the primitive. 
During optimization, small changes in the scale of the Gaussians result in changes of all the values of the texture. This can induce local minima in the optimization process that are visible as texture stretching, see Fig.~\ref{fig:stretch} (top).
Instead, we define a \emph{content-aware} representation in which textures are adapted to scene content. 
Our goal is to have textures that can faithfully represent the frequency of image details at the highest resolution present in the input images.
To do this, we fix the texel size with respect to the optimization process such that no optimizable parameters change it. We achieve this with a small modification to Eq.~\ref{eq:basic-u}:
\begin{equation}
    \mvec{u} = \frac{\mvec{p}^l}{k_i} + T_\text{offset}
\end{equation}
\noindent
where $k_i$ is the texel size and $T_\text{offset}$ is an offset to center the texture map.
\revision{rev9584 what does world units mean}{
In constrast to Eq.~\ref{eq:basic-u},
the intersection point $\mvec{p}^l$ and hence the texel size $k_i$ are defined in world space,
with the same units as the primitive's scales $\mvec{\sigma}$.
}

\revision{rev2866 texture resolution unclear}{This mapping implies that the textures do not have a fixed texture resolution.
On the contrary,
as the primitives grow or shrink,
more or less texture resolution is needed to cover their surface.
This is demonstrated in Fig.~\ref{fig:stretch} (bottom).
}
We modified our optimization routines to allocate and de-allocate texture resolution dynamically. 


Since the number of texels can grow quadratically, we risk running out of resources if every primitive  has a texture. Our goal is to maintain an expressive representation while carefully managing resources. To achieve this, we enforce two properties on our representation: First, the projected sampling frequency of the texture needs to be bounded by the image sampling frequency of the closest camera. This means that the projected texel size should never be smaller than the smallest input pixel that can see it. Second, texel sizes should adapt to \emph{scene content}, i.e., smaller texel sizes should be allocated for high frequency image content.

The first goal can be achieved using a conservative choice to determine minimum texel size as the pixel size back-projected in world space from the input training view closest to the primitive's center $k^p_{\mathrm{min}}$, similarly to \cite{reduced3dgs}.

Regarding the second goal,
 optimal texel sizes cannot  be determined at initialization
since pritimives change size and rotate dynamically during optimization.
We need a way to adapt to the freqency content of the scene progressively. 
To do this, we next introduce an adaptive strategy to determine texel size during optimization, where we downscale and upscale texture by adapting to the scene \emph{content}. 
We complement this approach with a resolution-aware primitive management method that add primitives to represent geometric -- rather than appearance -- error.
We describe these two components in the following sections.

\subsection{Progressive Adaptive Texel Size Determination}



Our goal is to adapt textures to the frequency content of the input images. We also need to achieve this in a progressive manner that fits well with the optimization process. We define a \emph{texel size-to-pixel} size ratio ${t_{2}p}_r$, which we manipulate during optimization using downscaling and upscaling operations.

We define the texel size $k$ based on the minimum pixel size ${k}^p_{\mathrm{min}}$, with ${t_{2}p}_r$ as follows:
\begin{equation}
    k = {k}^p_{\mathrm{min}} \cdot {t_{2}p}_r
\end{equation}

\noindent
We use this parameter to assign a low $t_2p_r$ to primitives in regions with fine details.
As a result, such primitives will have more texels for a given primitive size.
Similarly, primitives that lie in areas with low-frequency appearance will have a higher ratio,
giving them relatively fewer texels.
To provide a lower bound of 1 and facilitate texture rescaling operations,
$t_2p_r$ can only take values that are powers of two.

To this end, we propose two strategies to increase and decrease $t_2p_r$
leading to a downscale and an upscale of the texture map respectively.
Note that texel size and texture resolution are linked, but not the same. In the operations below, when texel size changes, texture resolution will change but only because the \emph{primitive size is unchanged}.

\begin{figure}[!h]
	\includegraphics[width=\linewidth]{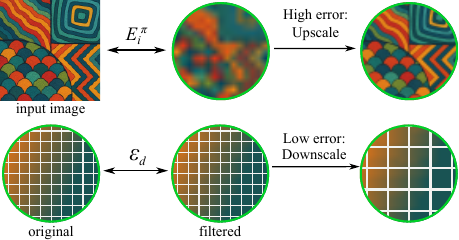}
\caption{
\label{fig:down-up}
We illustrate the downscale and upscale process used.
}
\end{figure}

\textbf{Increase of texel size-to-pixel size ratio - Downscaling}:
Our goal is to find texture maps that represent low frequency details and decrease their texel size accordingly.
We do this by applying a low-pass filter on textures and then comparing them to the originals.
The error $\mathcal{E}_d$ is weighted by the Gaussian falloff of the primitive,
leading to the following equation:
\begin{equation}
    \mathcal{E}_d = \frac{1}{\sum_\mvec{p}G(\mvec{p})}\sum_\mvec{p}G(\mvec{p})(T_\mathrm{orig}(\mvec{p}) - T_\mathrm{lowpass}(\mvec{p}))
\end{equation}

If this error is less than a threshold $\tau_\text{ds}$,
we assume that the texture map can be reconstructed with good-enough fidelity by the downscaled version.
Hence, we double  ${t_{2}p}_r$, which
reduces the total number of texture parameters by 4.
This reduces the resolution of texture, since the world-space size of the texels has grown. The size of the primitive is however unchanged. 

\textbf{Decrease of texel size-to-pixel size ratio - Upscaling}:
In regions where the texel size of the primitives involved is too large to capture the fine, underlying details, we expect our images to be blurry, and thus induce large error.
To identify these regions, we estimate primitive error using the approach presented in~\cite{revising}.
Specifically, for a primitive $i$, a ray $\mvec{r}$, a camera view $\pi$ and corresponding RGB error image $\mathcal{E}_\pi$,
we compute the per primitive error for a single view:
\begin{equation}
    E^\pi_i = \sum_{\mvec{r} \in \mathcal{P}_i} \mathcal{E}_\pi(\mvec{r})w^\pi_i(\mvec{r})
\end{equation}
where $\mathcal{P}_i$ are all the pixels covered by the primitive.
Here we assume that we have one ray per pixel,
emanating from its center.
Differently from \cite{revising},
instead of taking the maximum error over views, 
we perform a weighted sum,
with the weight being the total contribution of a primitive in that image.
\begin{equation}
    E_i = \frac{\sum_{\mvec{\pi} \in \Pi} E^\pi_i \overline{w_i^\pi}}{\sum_{\mvec{\pi} \in \Pi} \overline{w_i^\pi}}, \quad
    \overline{w_i^\pi} = \sum_{\mvec{r} \in \mathcal{P}_i}w_i^\pi(\mvec{r})
\end{equation}

\noindent
where $\Pi$ is the set of all input views.
This choice assigns an error value to each primitive that is proportional to their contribution to the rendered image.

We choose the top 10\% of primitives with the highest error, and upscale their textures by a factor of 4,
by halving $t_2p_r$.
With smaller sized texels,
these primitives can better fit to the scene content,
reducing their error.

\textbf{Progressive texel size adaptation.}
The calculation of the per primitive error $E_i$ and
the application of upscale/downscale process happens regularly during optimization.
Texel size adapts to scene content during this process, getting bigger for primitives that lie in low-frequency regions, and smaller for primitives that exhibit large error,
as shown in Fig.~\ref{fig:teaser}(right).

There are two main sources of error: appearance and geometry. For the first case, our upscaling approach reduces error as the optimization progresses.
However, for the case of geometry that is not well represented, an additional step is required.
To address this and complete our content-aware method, we next introduce resolution-aware splitting.

\subsection{Resolution-aware Primitive Management}

In Gaussian splatting methods where each primitive contains one color sample, \emph{densification} -- i.e., adding new primitives through cloning and splitting -- plays an essential role in improving the reconstruction of the scene.
This results in a local increase in both the geometric (position, scale, rotation) and appearance parameters (SH, opacity) that are treated together since they are tied to a single Gaussian primitive. 
However, this is not always ideal, since there are cases where the geometry is low frequency and has been approximated well, but we are missing texture detail, or conversely, cases where the color information is low frequency, but the underlying surface is not correctly represented. The latter case can appears as holes, ``softened'' edges or elongated primitives in places where they are not required. 

\begin{figure}[!h]
\includegraphics[width=\linewidth]{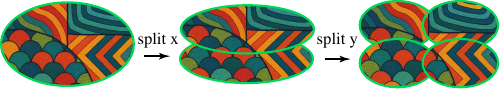} 
\caption{
\label{fig:splitting}
Left to right:The primitive has been upscaled to a resolution greater than $\tau_{tr}$ in both axes. Our splitting approach creates four new primitives, each with half the scale and texture resolution in both axes.}
\end{figure}

Our separation of geometric and appearance parameters provides an additional degree of freedom, allowing us to independently decide whether to increase the number of parameters linked to \emph{geometry}, i.e., increasing the number of primitives or to \emph{appearance}, i.e., increasing texture resolution.

Given that we do not have a supervision signal explicitly for geometry, 
we attempt to first match the appearance, using the upscaling approach described above.
If the error is still high despite upscaling,
we interpret the remaining error as geometric.
In these cases,
we can add more primitives,
locally increasing the geometric degrees of freedom.
\revision{rev2866 3dgs-mcmc}{
We found that densification strategies that were based on cloning either partially \cite{3dgs} or entirely in the case of 3DGS-MCMC \cite{3dgs-mcmc}, are incompatible with our representation. This is because
the superimposition of multiple textured primitives makes convergence difficult and requires an excessive number of parameters.
As a result our primitive management is based on \emph{splitting}, which fits well with our method. We describe this next.
}

\begin{figure}[!h]
\includegraphics[width=\linewidth]{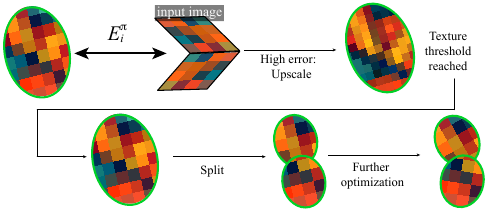} 
\caption{
\label{fig:densify}
The primitive has high error, and successive upscalings result in a texture resolution above threshold. In this case, the error is geometric rather than due to appearance. Our algorithm performs a split, and further optimization will match the geometry.}
\end{figure}

Similar to the approach for upscaling, we take the top 10\% of primitives with the highest error, and check which of these exceed a threshold $\tau_{\mathrm{tr}}$ for texture resolution. The primitive is replaced by two new primitives, each displaced by $\pm 1$ standard deviation of the Gaussian.
This process is performed separately for each axis; the new primitives have half the scale (size) in each corresponding axis,
as a result, half the texture resolution,
$G(1)$ times the opacity, where $G$ is the Gaussian.
The texture map of the newly added primitives is created by sampling the original point at the respective locations.
Fig.~\ref{fig:splitting} illustrates the splitting process for a primitive that happens to have a texture resolution greater than $\tau_{\mathrm{tr}}$ in both axes. A flowchart of how upscaling and splitting interact is shown in Fig.~\ref{fig:densify}. Alg.~\ref{alg:routines} displays in high level how the two methods are integrated in code.

\begin{table*}[!ht]
\footnotesize
\setlength\tabcolsep{1.5pt}
\centering
\scalebox{0.85}{
\begin{tabular}{@{}>{\raggedright\arraybackslash}p{1.8cm}|>{\centering\arraybackslash}p{0.8cm}>{\centering\arraybackslash}p{0.8cm}>{\centering\arraybackslash}p{0.8cm}>{\centering\arraybackslash}p{0.8cm}>{\raggedleft\arraybackslash}p{0.8cm}>{\centering\arraybackslash}p{0.9cm}>{\centering\arraybackslash}p{0.55cm}|>{\centering\arraybackslash}p{0.8cm}>{\centering\arraybackslash}p{0.8cm}>{\centering\arraybackslash}p{0.8cm}>{\centering\arraybackslash}p{0.8cm}>{\raggedleft\arraybackslash}p{0.8cm}>{\centering\arraybackslash}p{0.9cm}>{\centering\arraybackslash}p{0.55cm}|>{\centering\arraybackslash}p{0.8cm}>{\centering\arraybackslash}p{0.8cm}>{\centering\arraybackslash}p{0.8cm}>{\centering\arraybackslash}p{0.8cm}>{\raggedleft\arraybackslash}p{0.8cm}>{\centering\arraybackslash}p{0.9cm}>{\centering\arraybackslash}p{0.55cm}}
 & \multicolumn{7}{c|}{DeepBlending} & \multicolumn{7}{c|}{Mip-Nerf-360} & \multicolumn{7}{c}{Tanks\&Temples} \\
 & SSIM$\uparrow$ & PSNR$\uparrow$ & LPIPS$\downarrow$ & Points & Texels & Params & FPS & SSIM$\uparrow$ & PSNR$\uparrow$ & LPIPS$\downarrow$ & Points & Texels & Params & FPS & SSIM$\uparrow$ & PSNR$\uparrow$ & LPIPS$\downarrow$ & Points & Texels & Params & FPS \\
\midrule
3DGS-MCMC &  0.903 & 29.81 & 0.311 & 1323K & 0.0M & 78.1M & 265 & 0.831 & 27.82 & 0.232 & 2587K & 0.0M & 152.6M & 103 & 0.855 & 24.25 & 0.211 & 695K & 0.0M & 41.1M & 230 \\
\midrule
2DGS* & {\cellcolor{tabthird}} 0.899 & {\cellcolor{tabthird}} 29.52 & 0.324 & 1444K & 0.0M & {\cellcolor{tabsecond}} 83.8M & {\cellcolor{tabfirst}} 96 & {\cellcolor{tabsecond}} 0.801 & {\cellcolor{tabfirst}} 27.18 & {\cellcolor{tabthird}} 0.282 & 2079K & 0.0M & {\cellcolor{tabfirst}} 120.6M & {\cellcolor{tabfirst}} 91 & 0.834 & 23.37 & {\cellcolor{tabthird}} 0.239 & 872K & 0.0M & {\cellcolor{tabsecond}} 50.6M & {\cellcolor{tabfirst}} 165 \\
BBSplat & 0.898 & 29.25 & {\cellcolor{tabsecond}} 0.318 & 160K & 41.0M & 173.0M & {\cellcolor{tabthird}} 27 & 0.781 & 26.67 & {\cellcolor{tabsecond}} 0.273 & 237K & 60.9M & 257.0M & {\cellcolor{tabthird}} 23 & {\cellcolor{tabfirst}} 0.848 & {\cellcolor{tabfirst}} 23.62 & {\cellcolor{tabfirst}} 0.178 & 300K & 76.8M & 324.3M & {\cellcolor{tabthird}} 38 \\
GSTex & {\cellcolor{tabsecond}} 0.906 & {\cellcolor{tabsecond}} 29.63 & {\cellcolor{tabthird}} 0.323 & 1503K & 10.0M & {\cellcolor{tabthird}} 117.2M & 21 & {\cellcolor{tabfirst}} 0.802 & {\cellcolor{tabsecond}} 27.06 & 0.285 & 2025K & 10.0M & {\cellcolor{tabsecond}} 147.5M & 20 & {\cellcolor{tabsecond}} 0.842 & {\cellcolor{tabsecond}} 23.48 & 0.240 & 877K & 10.0M & {\cellcolor{tabthird}} 80.9M & 20 \\
Ours & {\cellcolor{tabfirst}} 0.907 & {\cellcolor{tabfirst}} 30.03 & {\cellcolor{tabfirst}} 0.303 & 222K & 21.6M & {\cellcolor{tabfirst}} 78.1M & {\cellcolor{tabsecond}} 70 & {\cellcolor{tabthird}} 0.795 & {\cellcolor{tabthird}} 27.00 & {\cellcolor{tabfirst}} 0.263 & 218K & 46.6M & {\cellcolor{tabthird}} 152.6M & {\cellcolor{tabsecond}} 67 & {\cellcolor{tabthird}} 0.835 & {\cellcolor{tabthird}} 23.43 & {\cellcolor{tabsecond}} 0.225 & 164K & 10.4M & {\cellcolor{tabfirst}} 41.1M & {\cellcolor{tabsecond}} 121 \\
\end{tabular}
}

\caption{\label{tab:default-eval}
We compare our method against 2DGS* trained with no geometric regularisations,
BBSplat and GSTex, with default settings.
We show standard quality metrics (PSNR, SSIM, L-PIPS), total number of primitives, texels and parameters.
Our method achieves competitive results while using significantly fewer, highly expressive primitives. \revision{}{We also include 3DGS-MCMC for completeness; please note that 3DGS-based methods achieve better NVS but worse geometry quality compared to all approaches based on 2DGS (see discussion Sec.~\ref{sec:eval}).} }

\end{table*}

\begin{table*}[!ht]
\footnotesize
\setlength\tabcolsep{2pt}
\centering
\scalebox{0.85}{
\begin{tabular}{@{}>{\raggedright\arraybackslash}p{1.1cm}|>{\centering\arraybackslash}p{0.8cm}>{\centering\arraybackslash}p{0.8cm}>{\centering\arraybackslash}p{0.8cm}>{\centering\arraybackslash}p{0.8cm}>{\raggedleft\arraybackslash}p{0.8cm}>{\centering\arraybackslash}p{0.9cm}>{\centering\arraybackslash}p{0.55cm}|>{\centering\arraybackslash}p{0.8cm}>{\centering\arraybackslash}p{0.8cm}>{\centering\arraybackslash}p{0.8cm}>{\centering\arraybackslash}p{0.8cm}>{\raggedleft\arraybackslash}p{0.8cm}>{\centering\arraybackslash}p{0.9cm}>{\centering\arraybackslash}p{0.55cm}|>{\centering\arraybackslash}p{0.8cm}>{\centering\arraybackslash}p{0.8cm}>{\centering\arraybackslash}p{0.8cm}>{\centering\arraybackslash}p{0.8cm}>{\raggedleft\arraybackslash}p{0.8cm}>{\centering\arraybackslash}p{0.9cm}>{\centering\arraybackslash}p{0.55cm}}
 & \multicolumn{7}{c|}{DeepBlending} & \multicolumn{7}{c|}{Mip-Nerf-360} & \multicolumn{7}{c}{Tanks\&Temples} \\
 & SSIM$\uparrow$ & PSNR$\uparrow$ & LPIPS$\downarrow$ & Points & Texels & Params & FPS & SSIM$\uparrow$ & PSNR$\uparrow$ & LPIPS$\downarrow$ & Points & Texels & Params & FPS & SSIM$\uparrow$ & PSNR$\uparrow$ & LPIPS$\downarrow$ & Points & Texels & Params & FPS \\
\midrule
BBSplat & {\cellcolor{tabthird}} 0.895 & {\cellcolor{tabsecond}} 28.93 & {\cellcolor{tabsecond}} 0.332 & 72K & 18.5M & 78.1M & {\cellcolor{tabthird}} 57 & {\cellcolor{tabsecond}} 0.768 & {\cellcolor{tabsecond}} 26.02 & {\cellcolor{tabsecond}} 0.291 & 141K & 36.1M & 152.6M & {\cellcolor{tabthird}} 33 & {\cellcolor{tabsecond}} 0.801 & {\cellcolor{tabsecond}} 22.77 & {\cellcolor{tabsecond}} 0.259 & 47K & 12.3M & 51.8M & {\cellcolor{tabsecond}} 87 \\
GSTex & {\cellcolor{tabsecond}} 0.896 & {\cellcolor{tabthird}} 28.29 & {\cellcolor{tabthird}} 0.354 & 222K & 21.6M & 78.1M & {\cellcolor{tabsecond}} 60 & {\cellcolor{tabthird}} 0.748 & {\cellcolor{tabthird}} 25.19 & {\cellcolor{tabthird}} 0.352 & 217K & 46.6M & 152.6M & {\cellcolor{tabsecond}} 46 & {\cellcolor{tabthird}} 0.745 & {\cellcolor{tabthird}} 20.65 & {\cellcolor{tabthird}} 0.363 & 164K & 10.4M & 41.1M & {\cellcolor{tabthird}} 67 \\
Ours & {\cellcolor{tabfirst}} 0.907 & {\cellcolor{tabfirst}} 30.03 & {\cellcolor{tabfirst}} 0.303 & 222K & 21.6M & 78.1M & {\cellcolor{tabfirst}} 70 & {\cellcolor{tabfirst}} 0.795 & {\cellcolor{tabfirst}} 27.00 & {\cellcolor{tabfirst}} 0.263 & 218K & 46.6M & 152.6M & {\cellcolor{tabfirst}} 67 & {\cellcolor{tabfirst}} 0.835 & {\cellcolor{tabfirst}} 23.43 & {\cellcolor{tabfirst}} 0.225 & 164K & 10.4M & 41.1M & {\cellcolor{tabfirst}} 121 \\
\end{tabular}
}

\caption{\label{tab:same-param-eval}
We compare against BBSplat and GSTex in a same parameter setting,
by adjusting their primitive count and texels accordingly.
The slight discrepancy in BBSplat's parameters in Tanks \& Temples is due to a sphere sampling method they use to model the background and distant objects.
}
\end{table*}

\subsection{Training and Regularisation}
As in \cite{3dgs},
we train our model using the weighted sum of the per-pixel $\mathcal{L}_1$
and the structural similarity losses,
between the ground truth and the rendered images:
\begin{equation}
    \mathcal{L}_{\text{RGB}} = (1-\lambda_{\text{SSIM}})\mathcal{L}_1 + \lambda_{\text{SSIM}}\mathcal{L}_{\text{SSIM}}
\end{equation}
with $\lambda_{\text{SSIM}}=0.2$.

We observed that textures could finish converging to high-frequency settings,
even though alpha-blending would create a smooth result.
This prevented them from being downscaled,
leading to high parameter usage.
To resolve this,
we apply a sparsity regularization on the texel values,
pushing them towards zero,
\begin{equation}
    \mathcal{L}_{\text{texture}} = \lambda_{\text{texture}} \sum_i |\mvec{c}_i^{\mathcal{T}_i}|
\end{equation}
We constrain the texel values in the $[-1,1]$ range,
using a sigmoid activation,
scaled and shifted accordingly:
\begin{equation}
    \mvec{c}^{\mathcal{T}_i} = 2\sigma(\mvec{c}'^{\mathcal{T}_i})-1
\end{equation}
where $\sigma(x)$ is the sigmoid function and $\mvec{c}'^{\mathcal{T}_i}$ are the unactivated texture features.
Intuitively,
this parametrization coupled with the sparsity loss
forces the view-dependent color $\mvec{SH}(\mvec{d})$ to learn the base color of the surface,
while the texels operate as offsets to model high-frequency details.
An example of this is illustrated in Fig.~\ref{fig:teaser}(left).

Finally,
as in \cite{reduced3dgs, 3dgs-mcmc},
we incentivize low-contribution primitives to effectively disappear by applying an opacity regularization term:
\begin{equation}
    \mathcal{L}_{\text{opacity}} = \lambda_{\text{opacity}} \frac{1}{N}\sum_i^N \mvec{o}_i.
\end{equation}

Taking both training objectives and regularizations into consideration,
the total loss is formed as:
\begin{equation}
    \mathcal{L} = \mathcal{L}_{\text{RGB}} + \mathcal{L}_{\text{texture}} + \mathcal{L}_{\text{opacity}}.
\end{equation}

\begin{figure*}[!h]
\centering
\setlength{\tabcolsep}{1.5pt}

    \newcommand{\resultfigwidth}{.193\textwidth}
	\begin{tabular}{cc|cccc}
		& Ground Truth & 2DGS* & BBSplat & GStex & Ours \\
        \raisebox{0.25\height}{\rotatebox{90}{\small Mip-Nerf360}} &
        \spyimage{red}{\resultfigwidth}{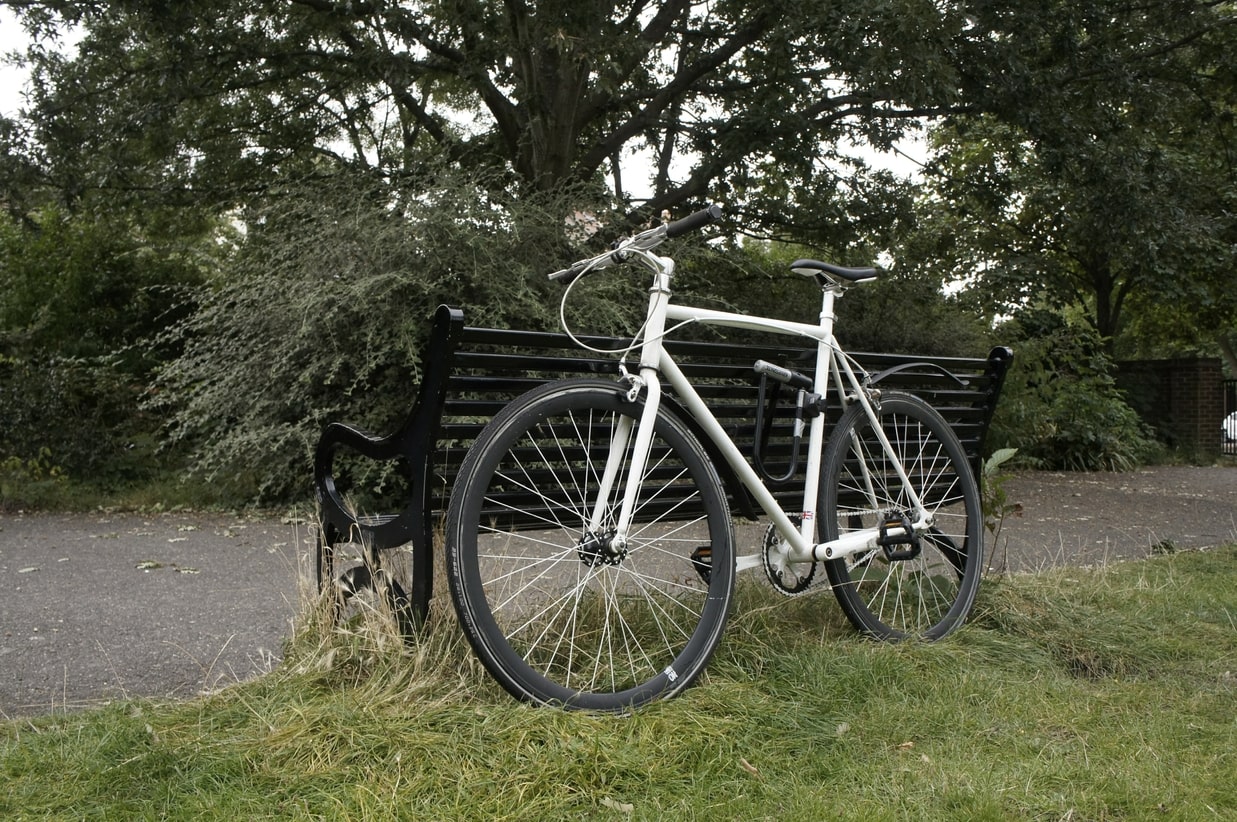}{0.4, 1.5}{1.1, 0.6} &
        \spyimage{red}{\resultfigwidth}{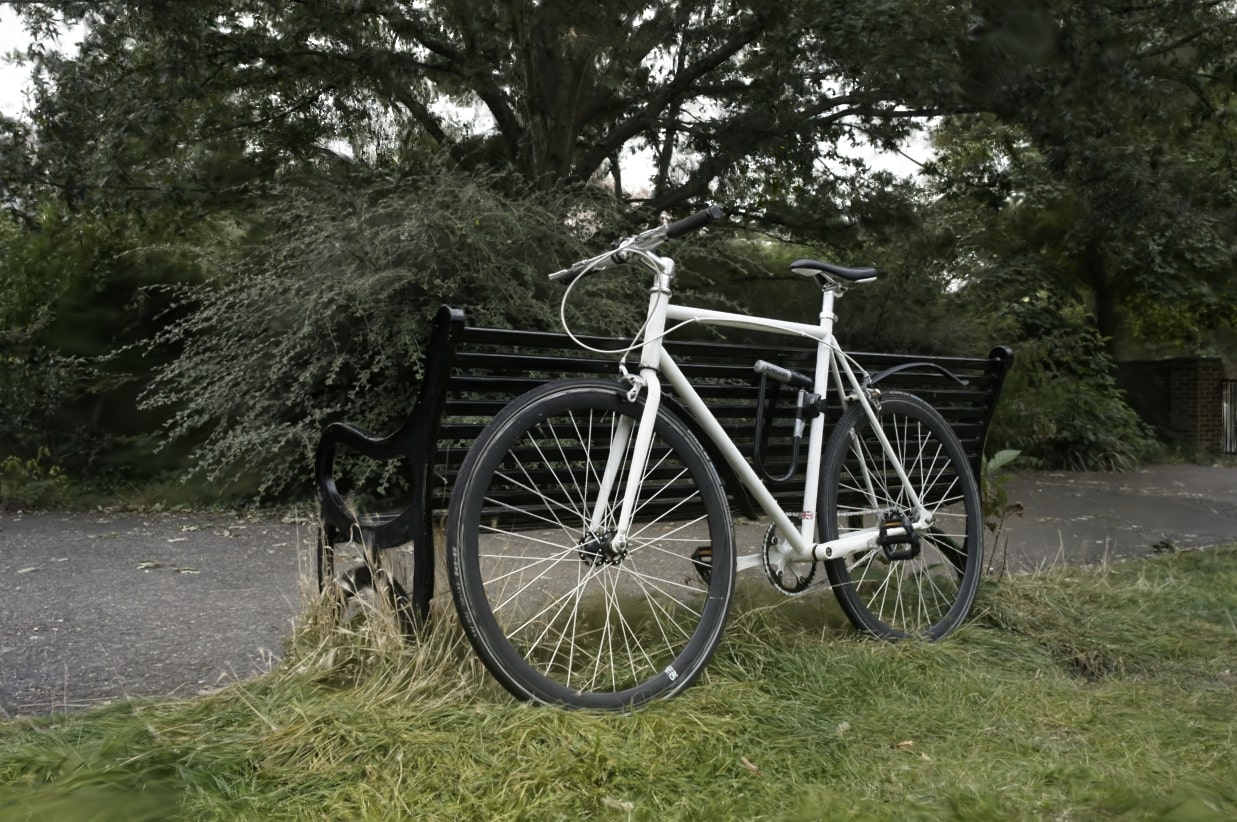}{0.4, 1.5}{1.1, 0.6} &
        \spyimage{red}{\resultfigwidth}{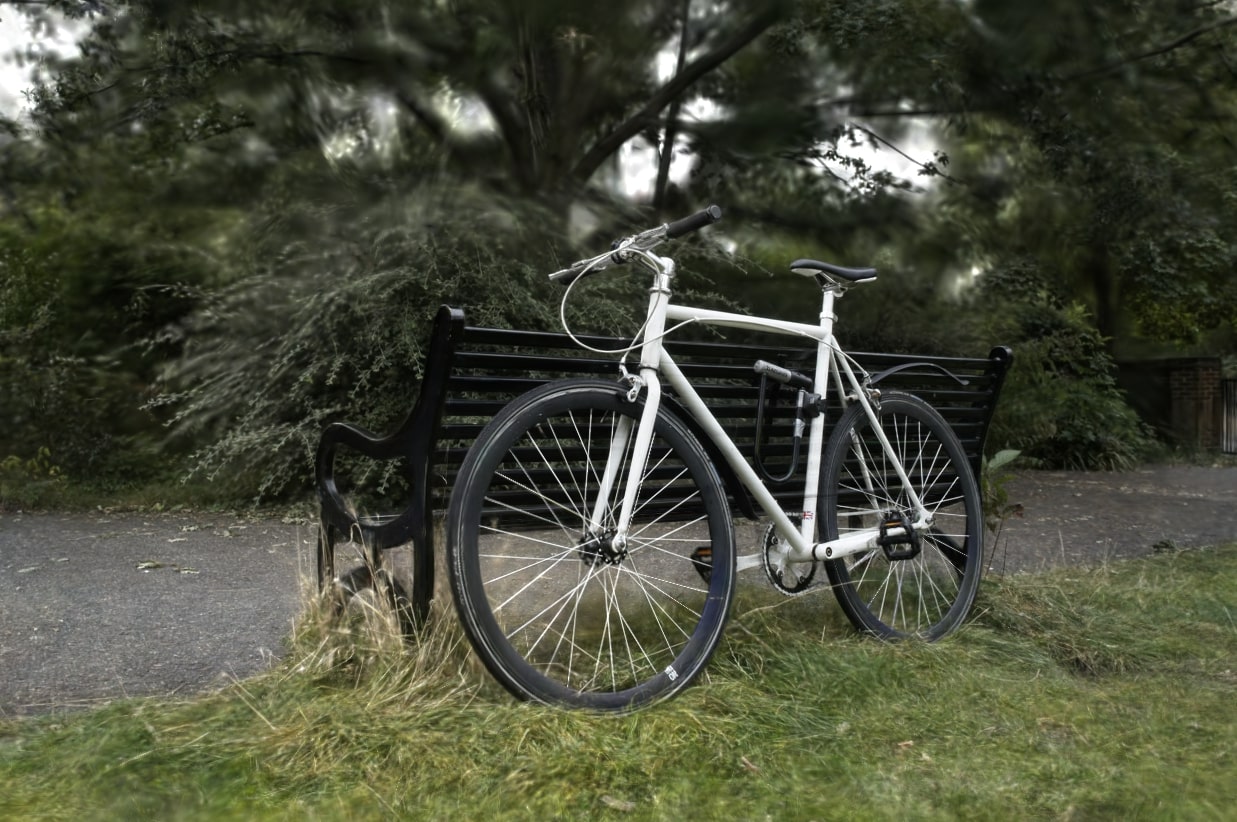}{0.4, 1.5}{1.1, 0.6} &
        \spyimage{red}{\resultfigwidth}{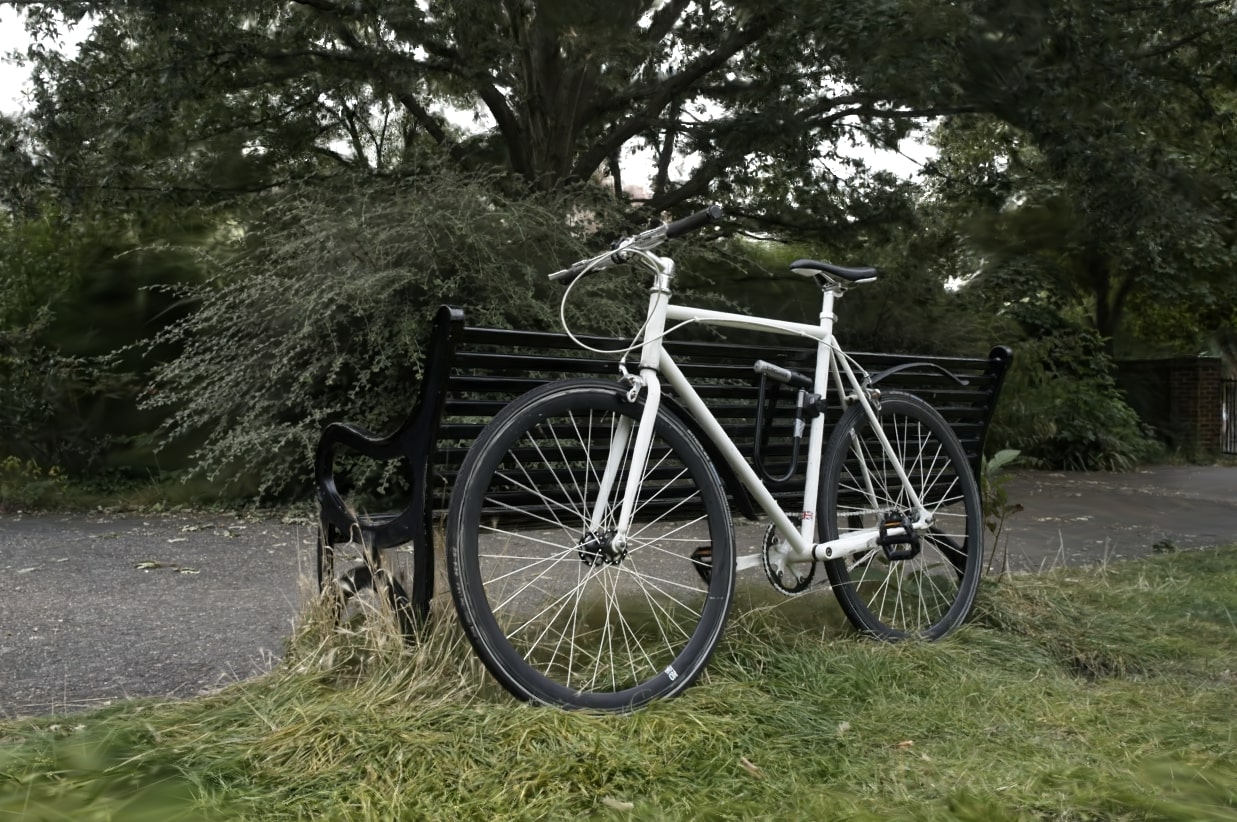}{0.4, 1.5}{1.1, 0.6} &
        \spyimage{red}{\resultfigwidth}{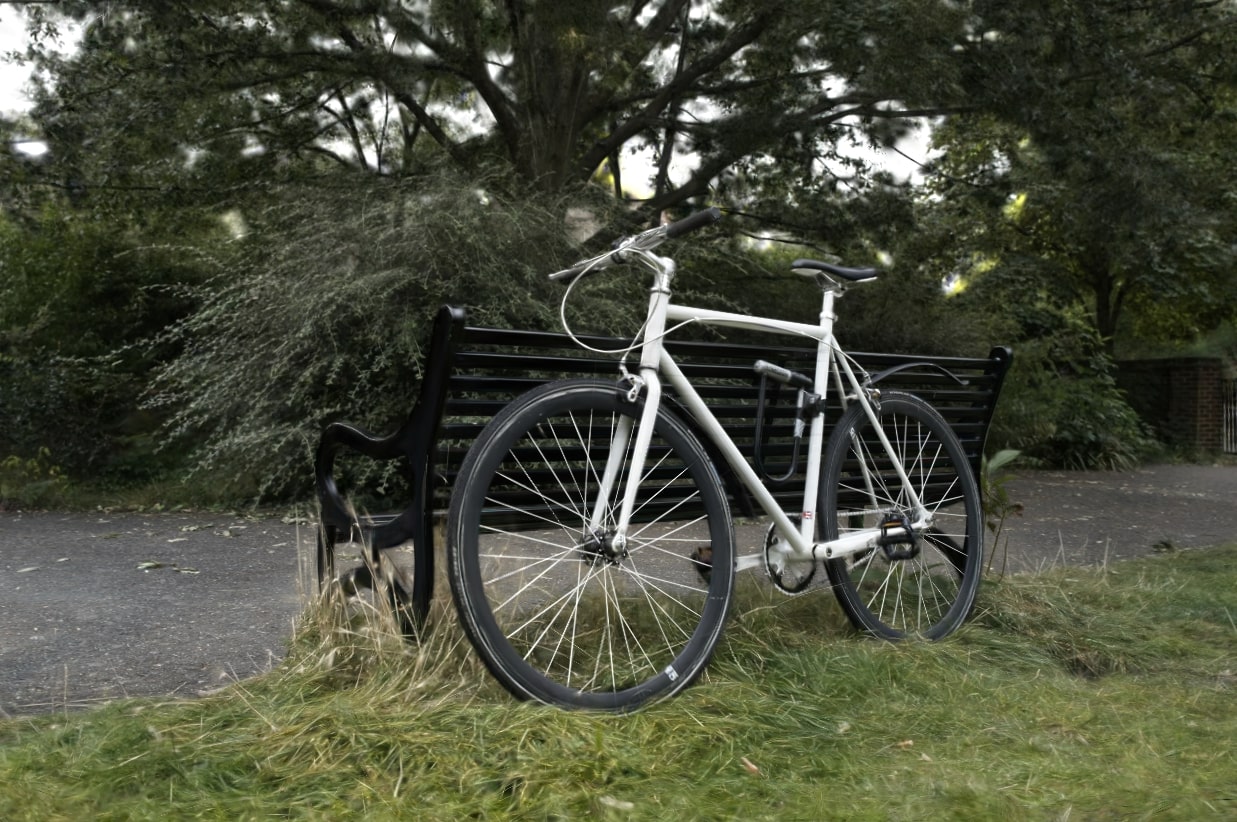}{0.4, 1.5}{1.1, 0.6} \\
		
        \raisebox{0.18\height}{\rotatebox{90}{\small Deep Blending}} &
        \spyimage{red}{\resultfigwidth}{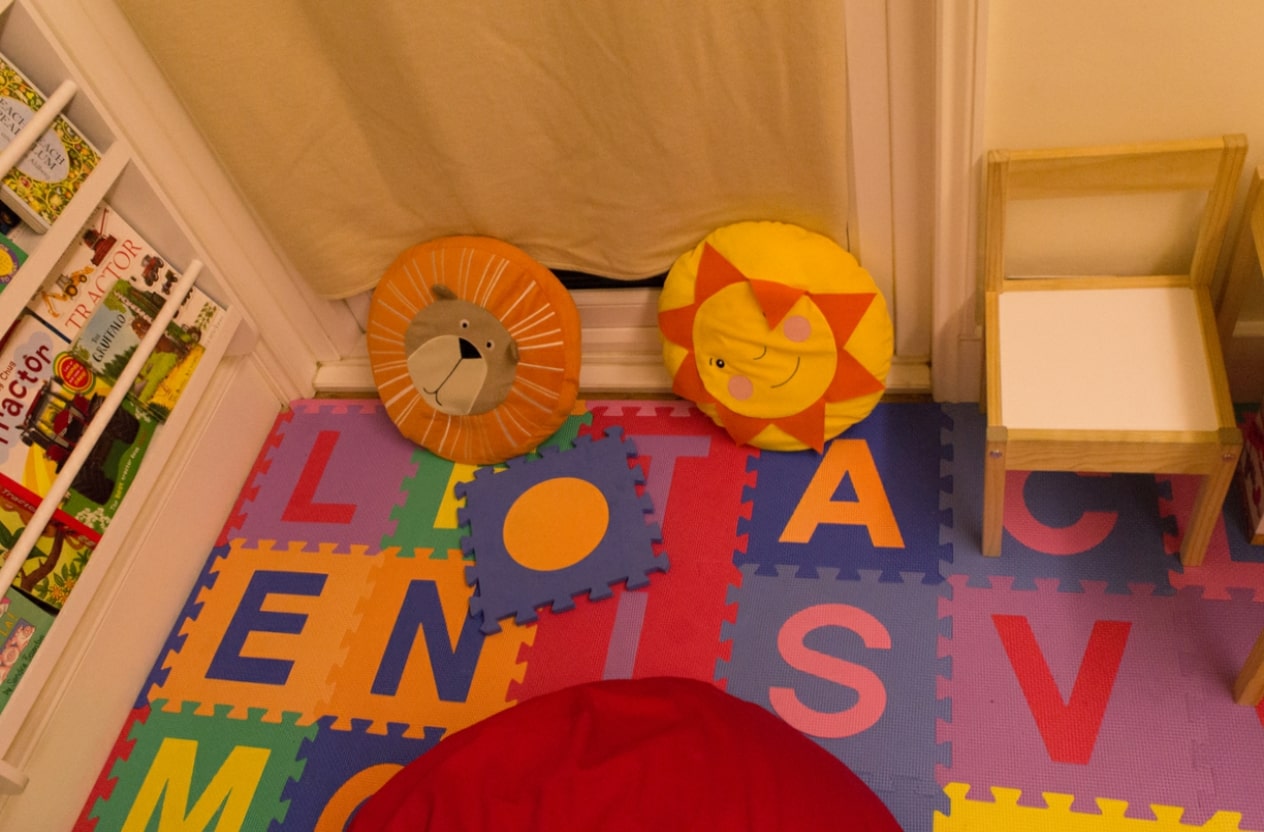}{3, 0.3}{3.3, 1.3} &
        \spyimage{red}{\resultfigwidth}{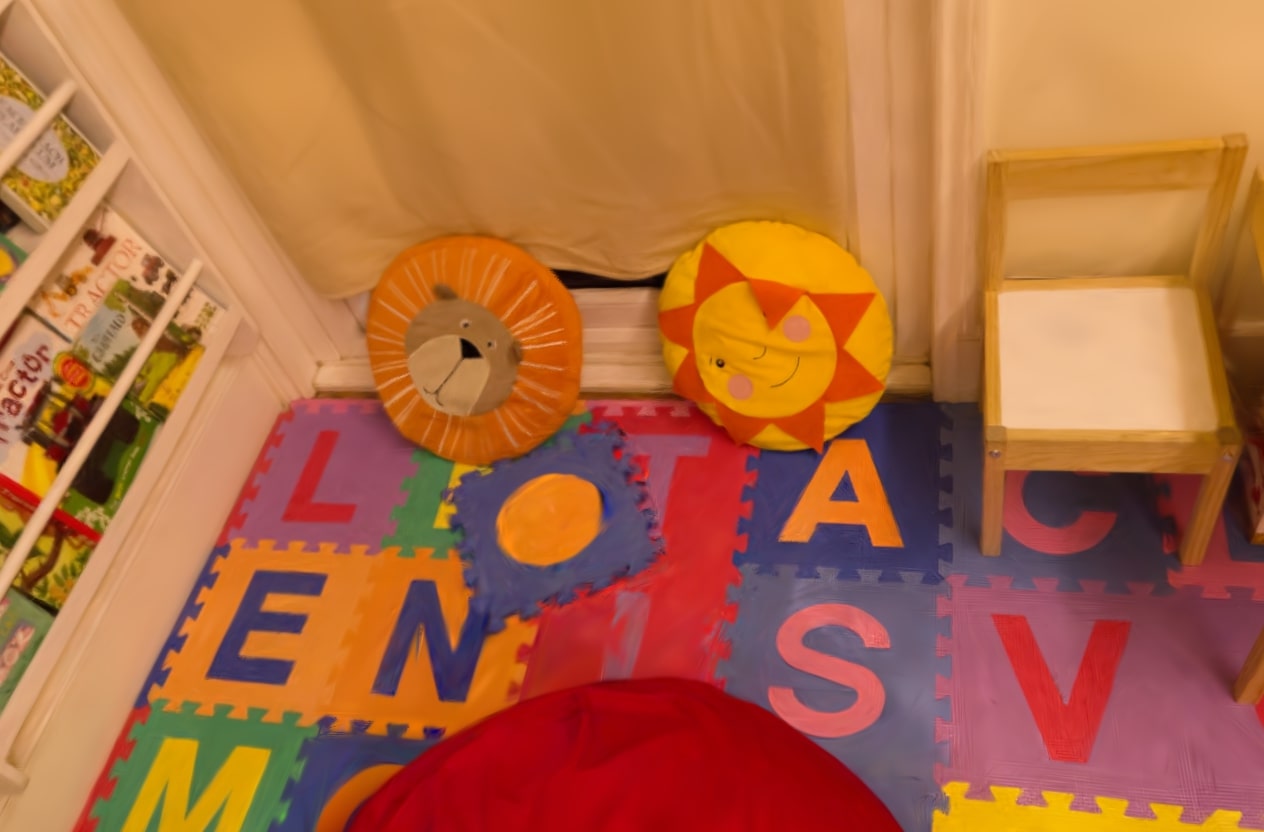}{3, 0.3}{3.3, 1.3} &
        \spyimage{red}{\resultfigwidth}{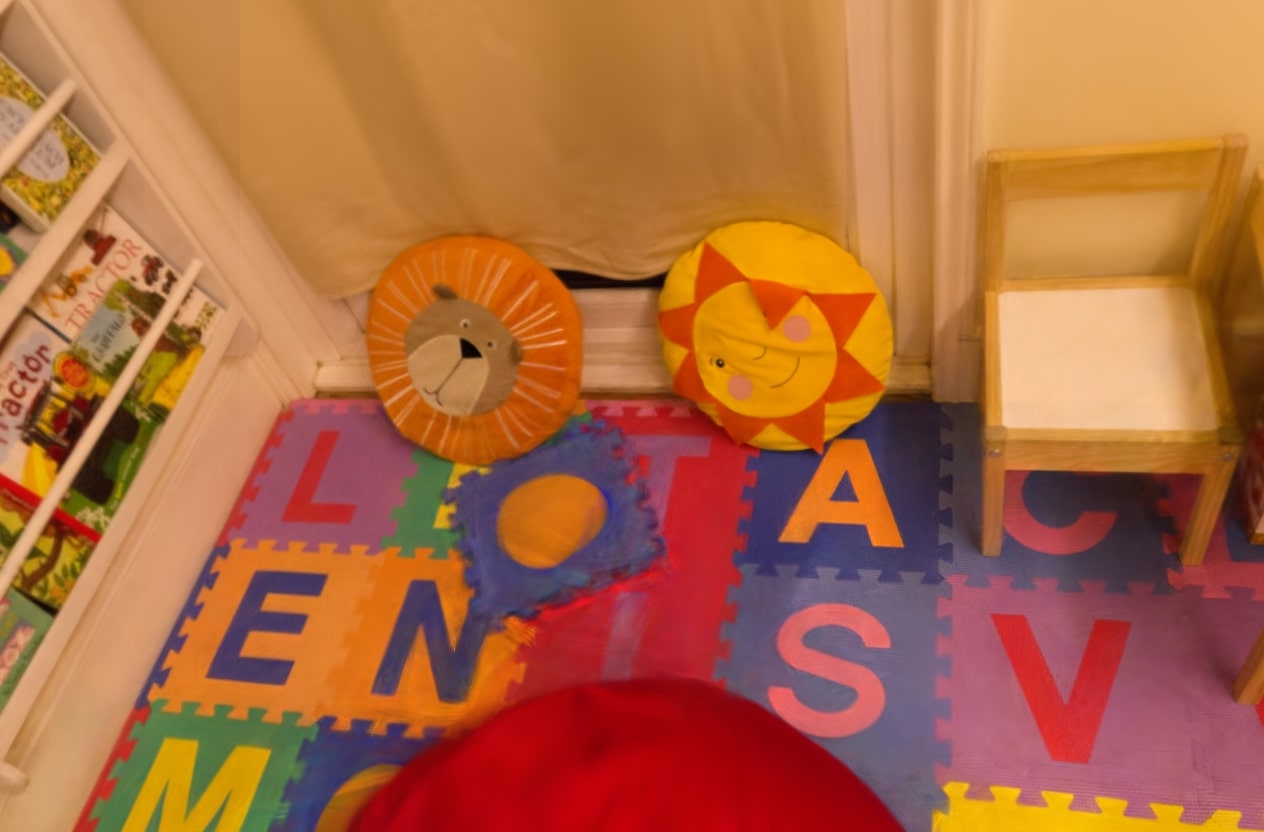}{3, 0.3}{3.3, 1.3} &
        \spyimage{red}{\resultfigwidth}{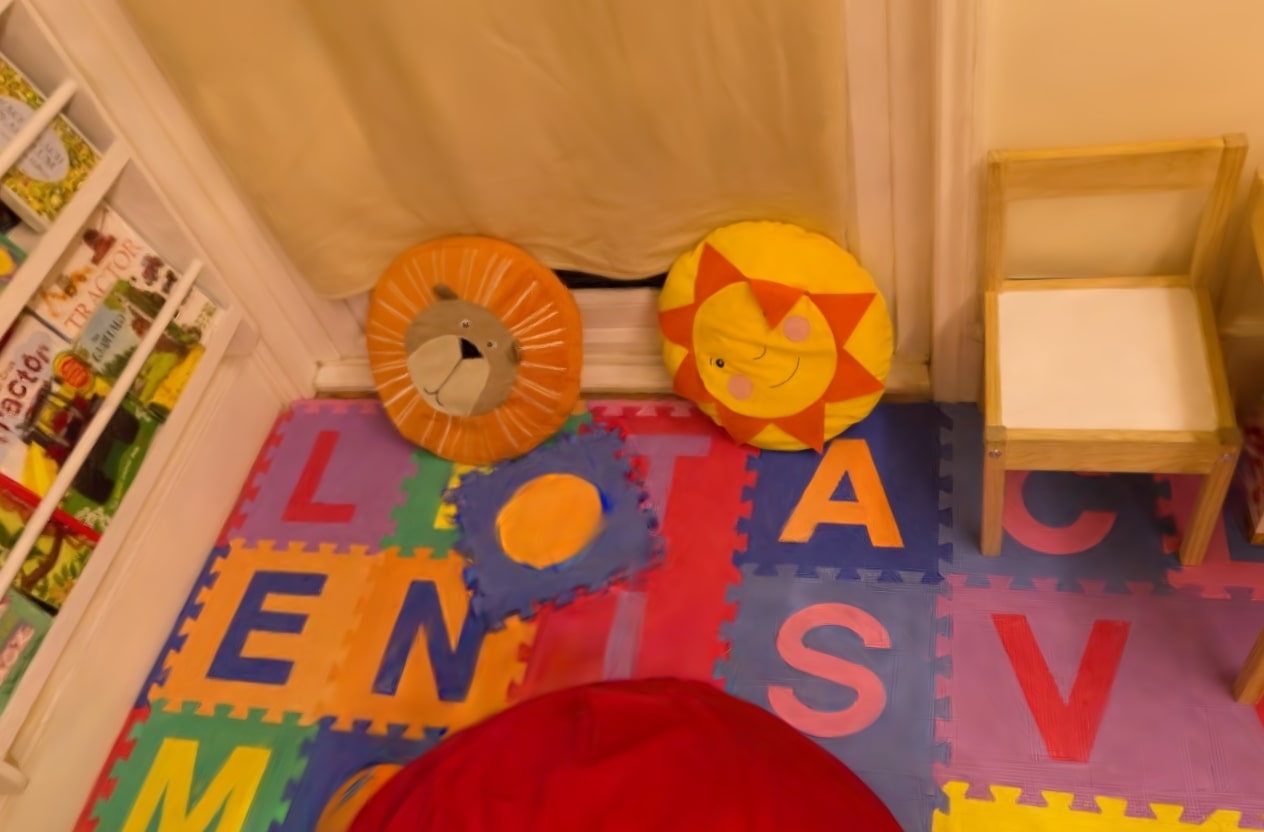}{3, 0.3}{3.3, 1.3} &
        \spyimage{red}{\resultfigwidth}{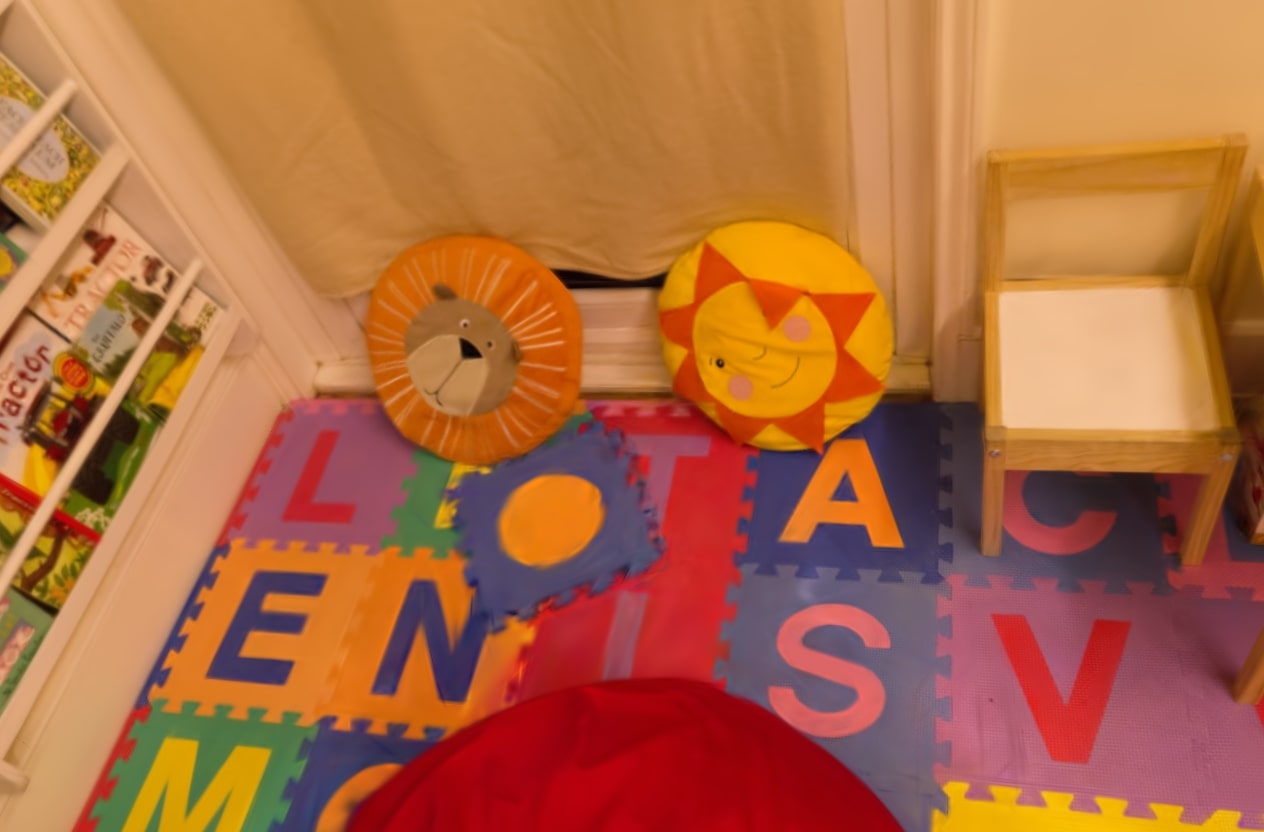}{3, 0.3}{3.3, 1.3} \\

        \rotatebox{90}{\small Tanks\&Temples} &
        \spyimage{red}{\resultfigwidth}{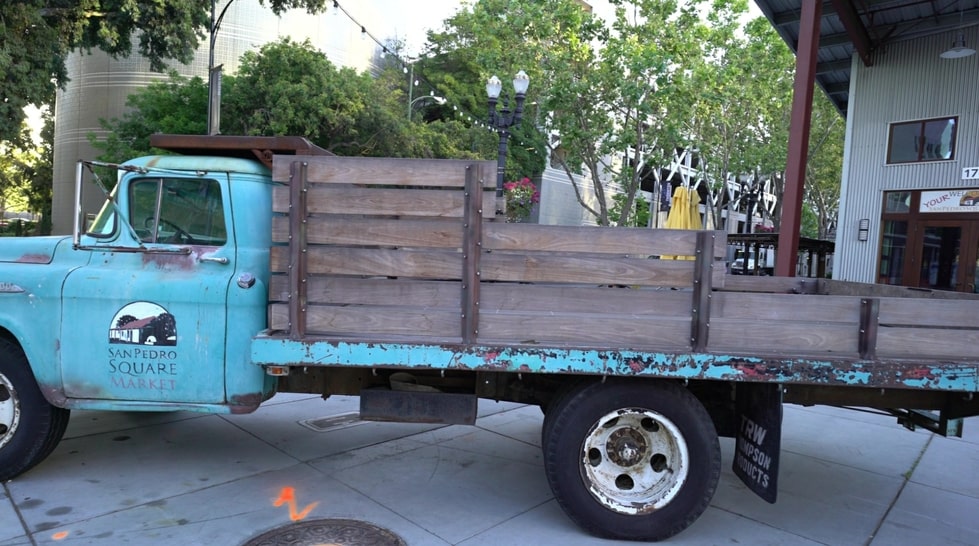}{3.1, 1.65}{3.1, 0.6} &
        \spyimage{red}{\resultfigwidth}{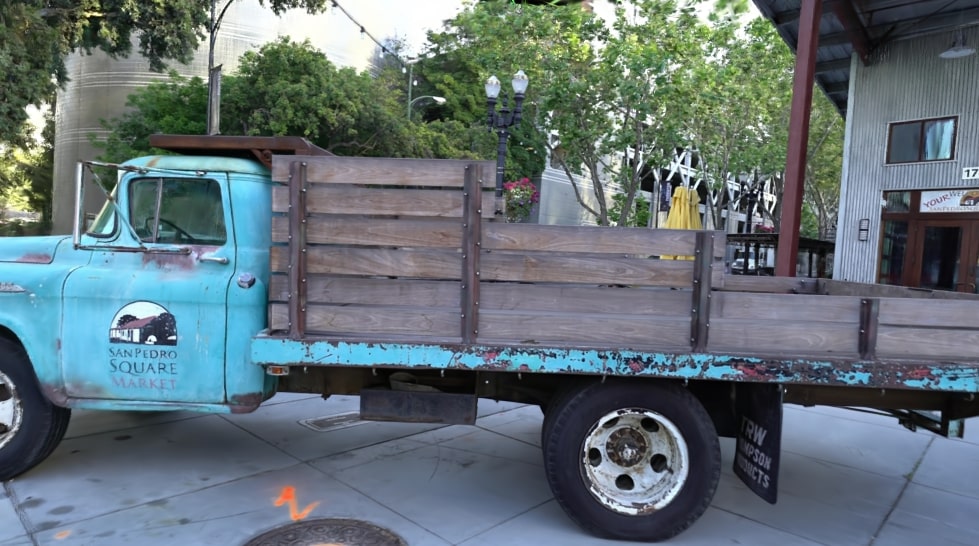}{3.1, 1.65}{3.1, 0.6} &
        \spyimage{red}{\resultfigwidth}{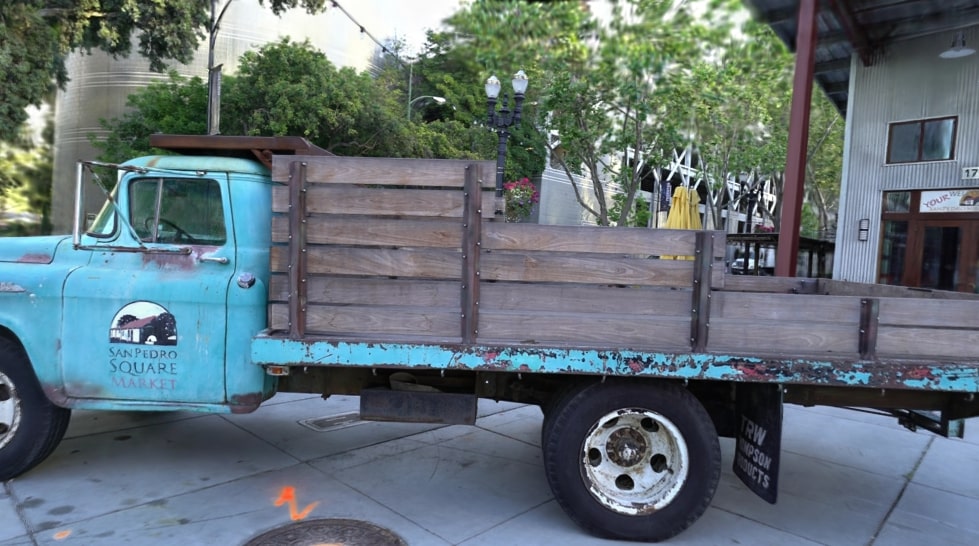}{3.1, 1.65}{3.1, 0.6} &
        \spyimage{red}{\resultfigwidth}{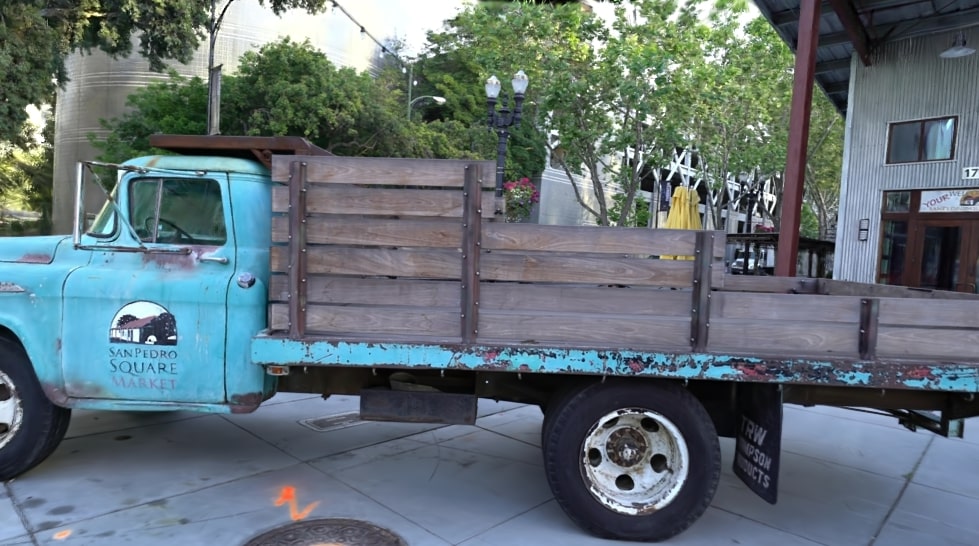}{3.1, 1.65}{3.1, 0.6} &
        \spyimage{red}{\resultfigwidth}{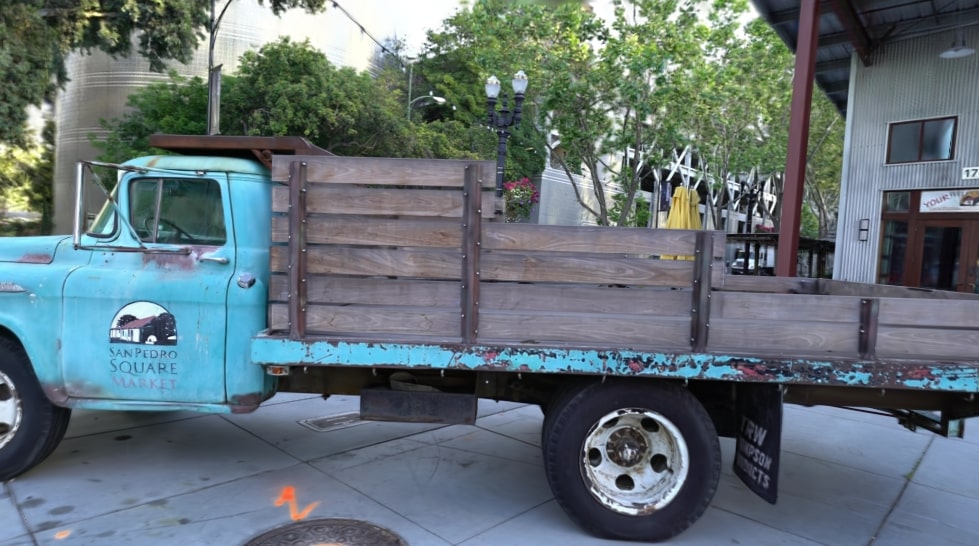}{3.1, 1.65}{3.1, 0.6} \\
	\end{tabular}
\caption{
\label{fig:default-eval}
We provide renderings from one scene per dataset for every method,
trained with default settings.
Our method is able to reconstruct high frequency details on images,
even while using fewer parameters compared to the other methods.
}
\end{figure*}

\section{Implementation}
We have implemented our method using the 3DGS codebase, introducing 2D primitives as in 2DGS. 
Unlike 2DGS,
we do not use the $\mathbb{R}^2 \rightarrow \mathbb{R}^2$ transformation to find the intersection point using three non-parallel planes.
Instead, we use Eq.~\ref{eq:intersection_point} directly in camera space.
We did not observe any instability issues.
\revision{rev6315 runtime performance}{
Since our method requires per-ray querying of texture maps,
 both the forward and backward passes incur additional overhead,
that is not compensated by the smaller number of primitives rendered.
In general, compared to 2DGS,
 training time is 1.5-2 times longer, and rendering is 25\% slower.
}

As the texture grids can have different resolutions,
we created custom ``jagged'' tensor data structure to be used for their storage.
The texture resolution for each primitive gets (de-)allocated dynamically,
so that it covers the extent of the primitive,
that is $\pm$3 standard deviations.
This dynamic memory management happens every 100 iterations.
A hard limit of 256 texture resolution is enforced to avoid using too much memory.
When a primitive grows outside of its allocated texture grid,
either because memory has not been allocated yet or it has surpassed the hard limit,
zero-padding is used. Since we are storing an offset, zero-padding reduces to the DC color.

We observed that limiting the $t_2p_r$ to be greater or equal to 2 did not have a detrimental effect on the visual quality,
as subtexel scaled details can be retrieved
thanks to alpha blending and the overlap of our primitives.
We thus allow upscaling to happen only for primitives with a $t_2p_r$ greater or equal to 2.



\revision{rev4160 memory/storage}
{
\textbf{Storage and Memory Considerations.}
The size of our representation is tied to the number of parameters used,
with each parameter saved as a 4-byte float,
as in previous Gaussian Splatting implementations.
However,
most Gaussian Splatting compression techniques (see \cite{3dgszip}) are 
applicable to our approach either directly or with minor modifications.
For our newly introduced texture maps,
our choice to use a sigmoid activation to limit the effective range of the texel values allows for a very efficient application of K-means clustering compression. Please see Tab.~\ref{tab:per-scene} for complete statistics.
}

\section{Results and Evaluation}

We test our method on 13 indoor and outdoor scenes from three different standard datasets.
We evaluate on all scenes of Mip-NeRF 360 \cite{mipnerf360},
two scenes from Tanks \& Temples \cite{tnt},
as well as two scenes from Deep Blending \cite{db}.
Fig.~\ref{fig:default-eval} shows that our method faithfully reconstructs the input images.

\subsection{Evaluation}
\label{sec:eval}

We compare our method against 2DGS \cite{2dgs} as a baseline,
and two 2DGS texturing methods,
namely (unpublished) BBSplat~\cite{bbsplat} and GStex \cite{gstex},
for which the code was available at the time of submission.
\revision{rev9584 2866 comparison to textured-3dgs 3dgs-mcmc}
{
Directly comparing to 3DGS-based models is not straightforward and might lead to misleading conclusions,
since these models have different properties and consistently perform better for NVS but worse in geometry reconstruction,
compared to 2DGS-based solutions~\cite{2dgs}.
For the same reason,
and because the code was not available at the time of submission,
we do not compare against \cite{textured3dgs},
which is a 3DGS-based method with textured primitives.
This model uses a pretrained 3DGS-MCMC model as initialization,
similar to GStex,
and includes RGBA textures,
similar to BBSplat.
We thus expect it to suffer from the disadvantages of both these two models (see discussion of the second experiment).
}
We ran these methods on all scenes using our machines to ensure fair comparison.
For 2DGS,
the normal consistency and depth distortion regularization terms were disabled,
as they are designed to enhance the geometric reconstruction
but often result in lower novel view synthesis (NVS) quality.

\begin{figure*}[!h]
\centering
\setlength{\tabcolsep}{1.5pt}

    \newcommand{\resultfigwidth}{.24\textwidth}

\begin{tabular}{cc|ccc}
    & Ground Truth & BBSplat & GStex & Ours \\
    \raisebox{0.4\height}{\rotatebox{90}{\small Mip-Nerf360}} &
    \spyimage{red}{\resultfigwidth}{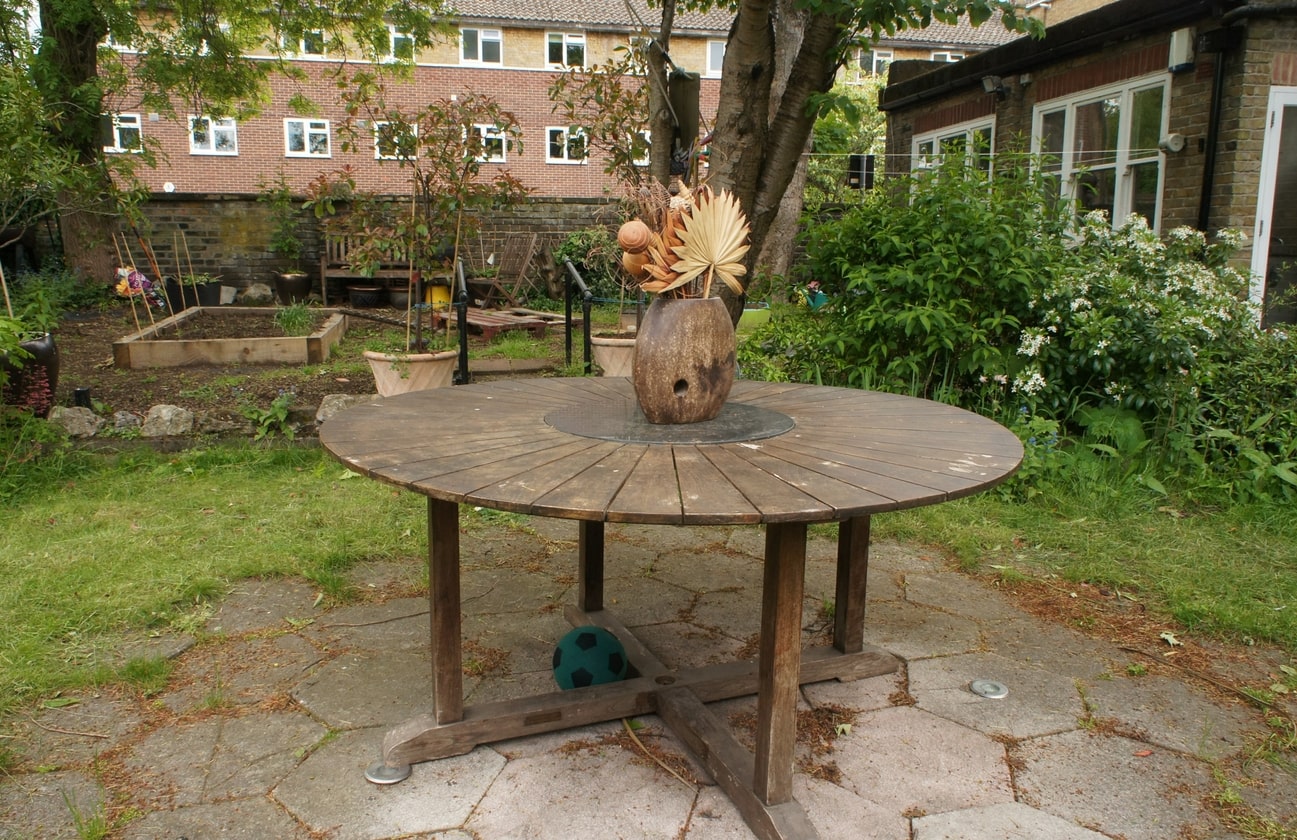}{1.4, 2.5}{1.1, 0.7} &
    \spyimage{red}{\resultfigwidth}{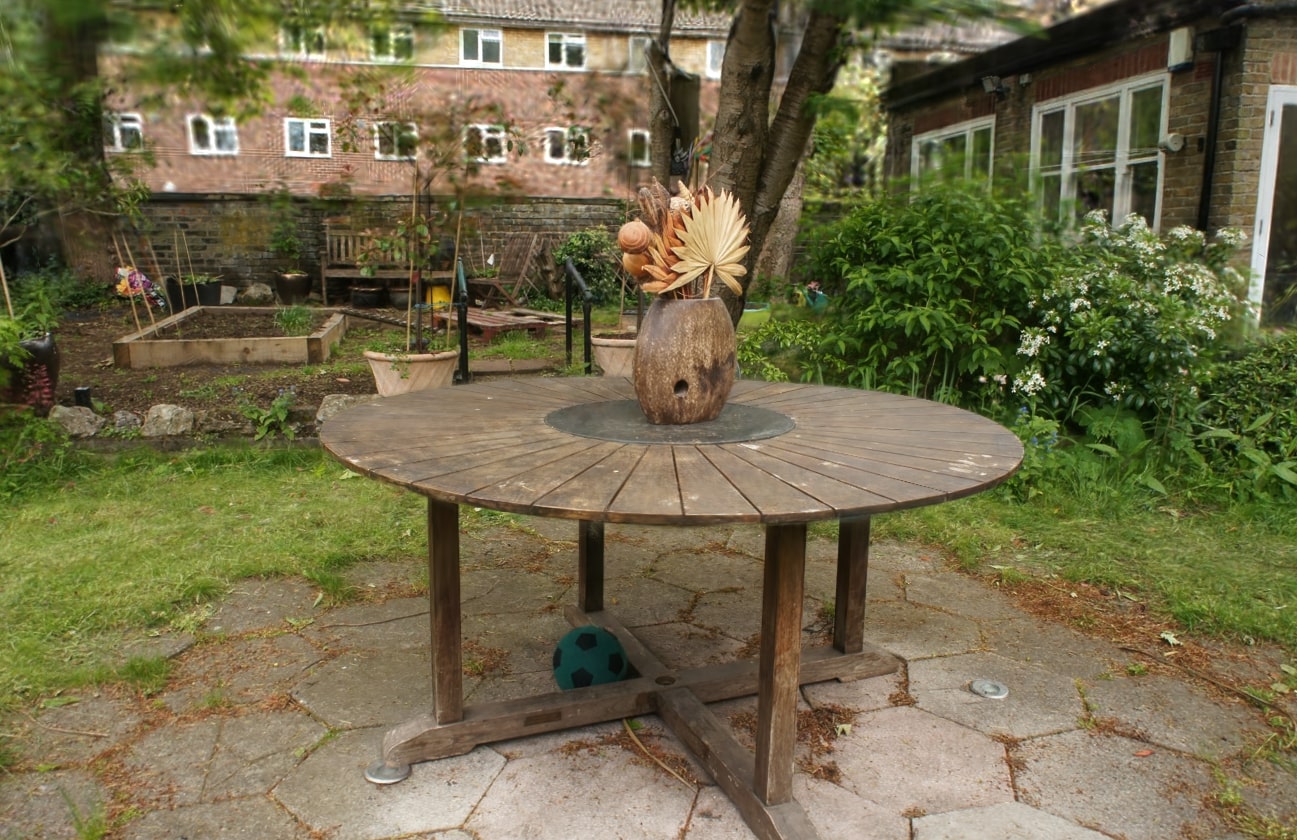}{1.4, 2.5}{1.1, 0.7} &
    \spyimage{red}{\resultfigwidth}{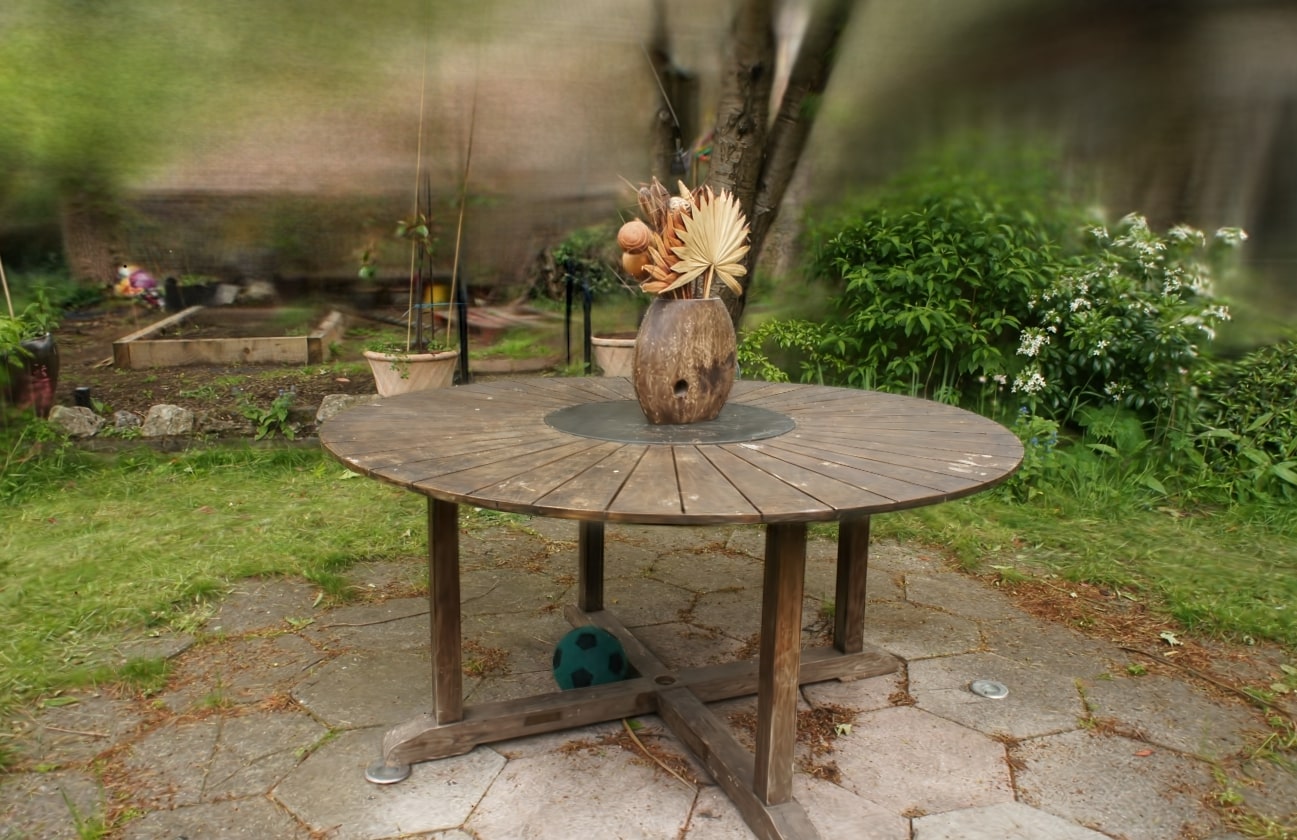}{1.4, 2.5}{1.1, 0.7} &
    \spyimage{red}{\resultfigwidth}{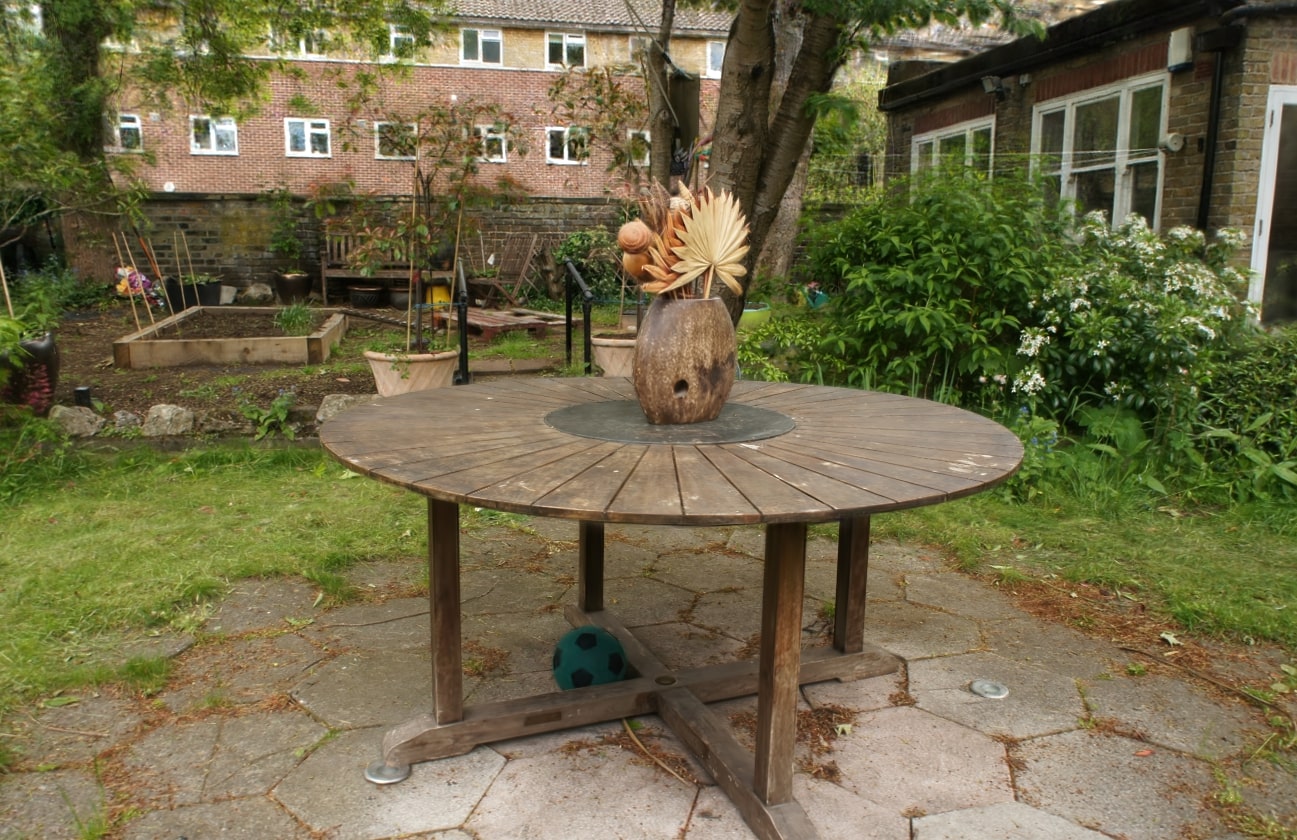}{1.4, 2.5}{1.1, 0.7} \\
    
    \raisebox{0.3\height}{\rotatebox{90}{\small Deep Blending}} &
    \spyimage{red}{\resultfigwidth}{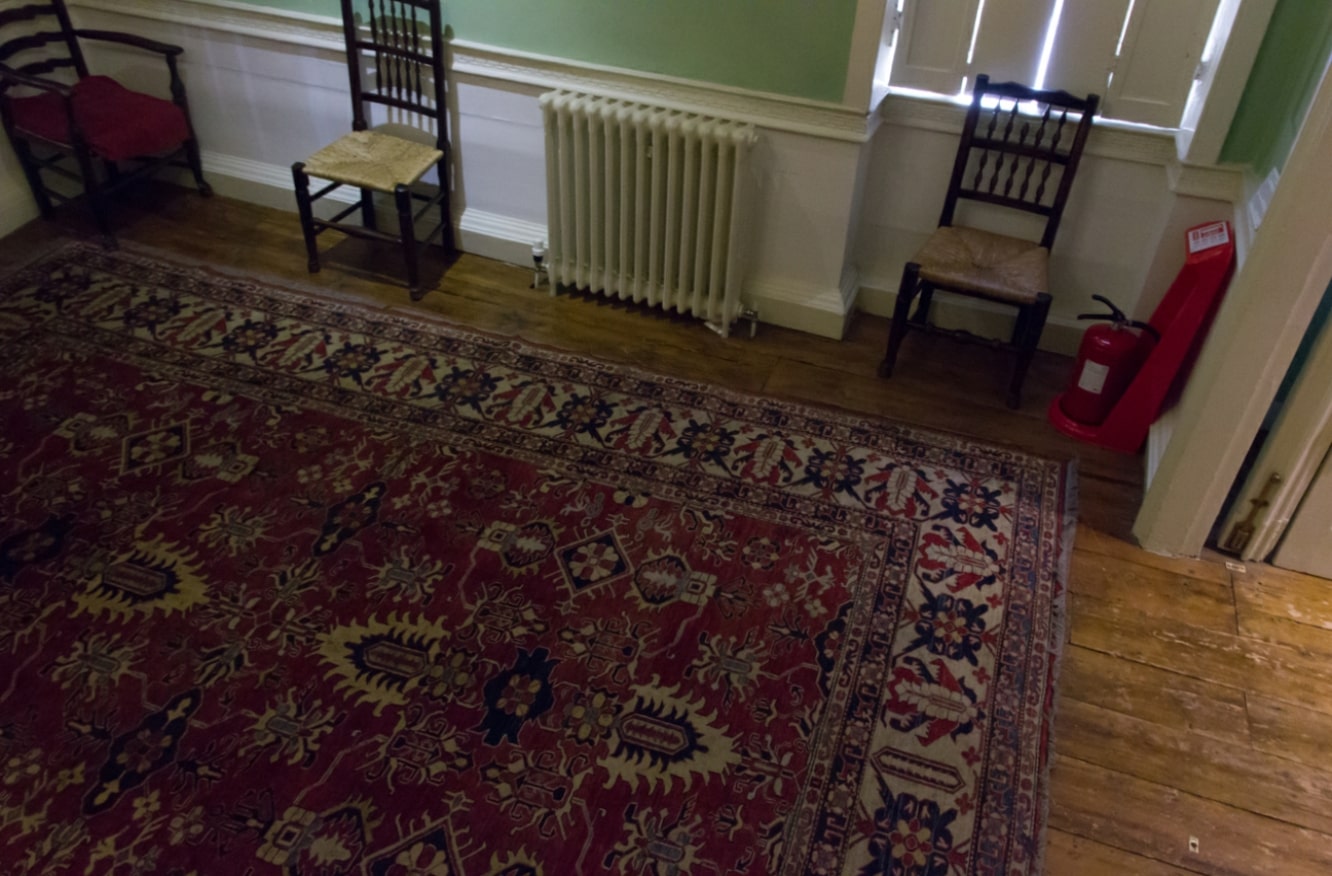}{2.5, 0.3}{4, 0.7} &
    \spyimage{red}{\resultfigwidth}{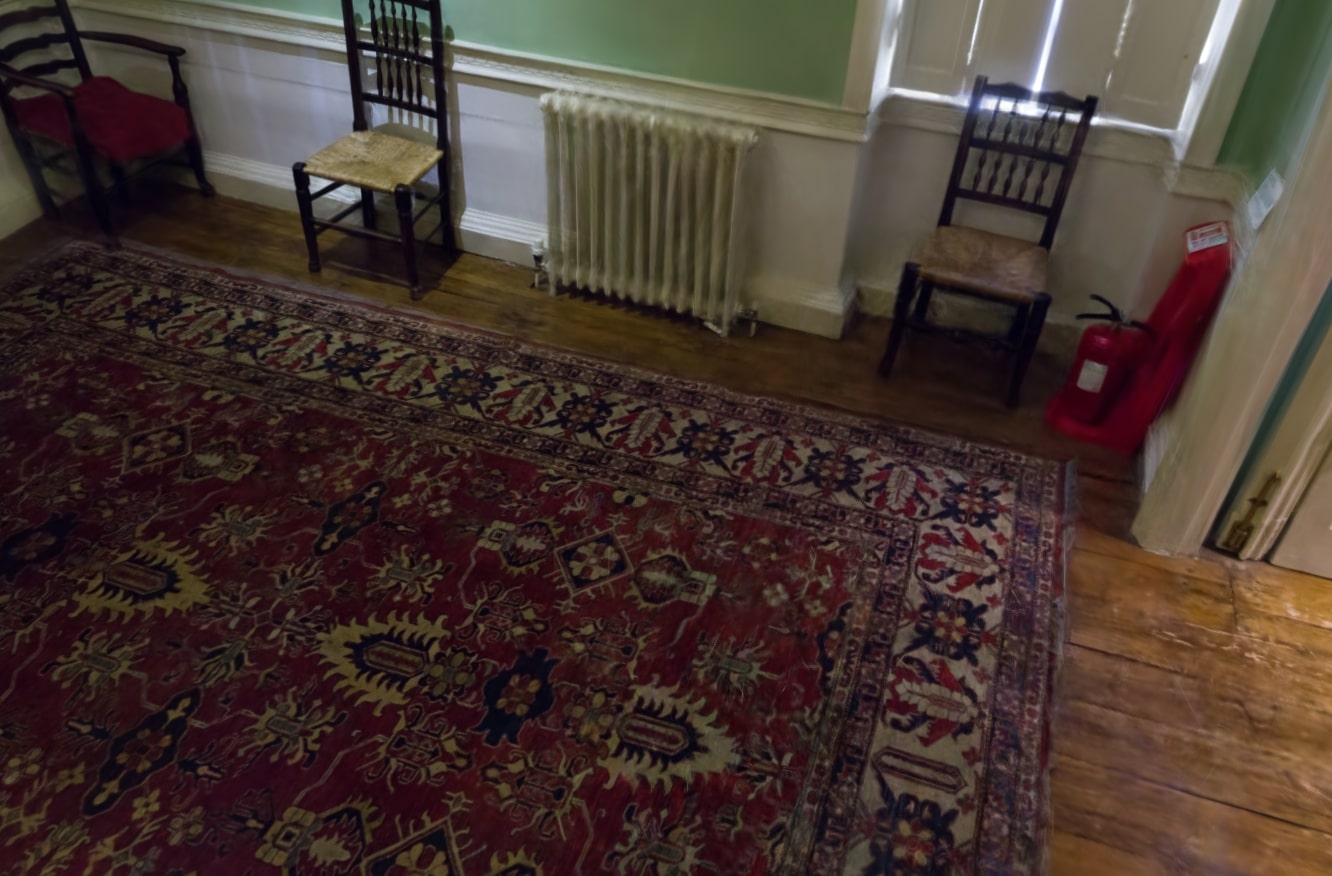}{2.5, 0.3}{4, 0.7} &
    \spyimage{red}{\resultfigwidth}{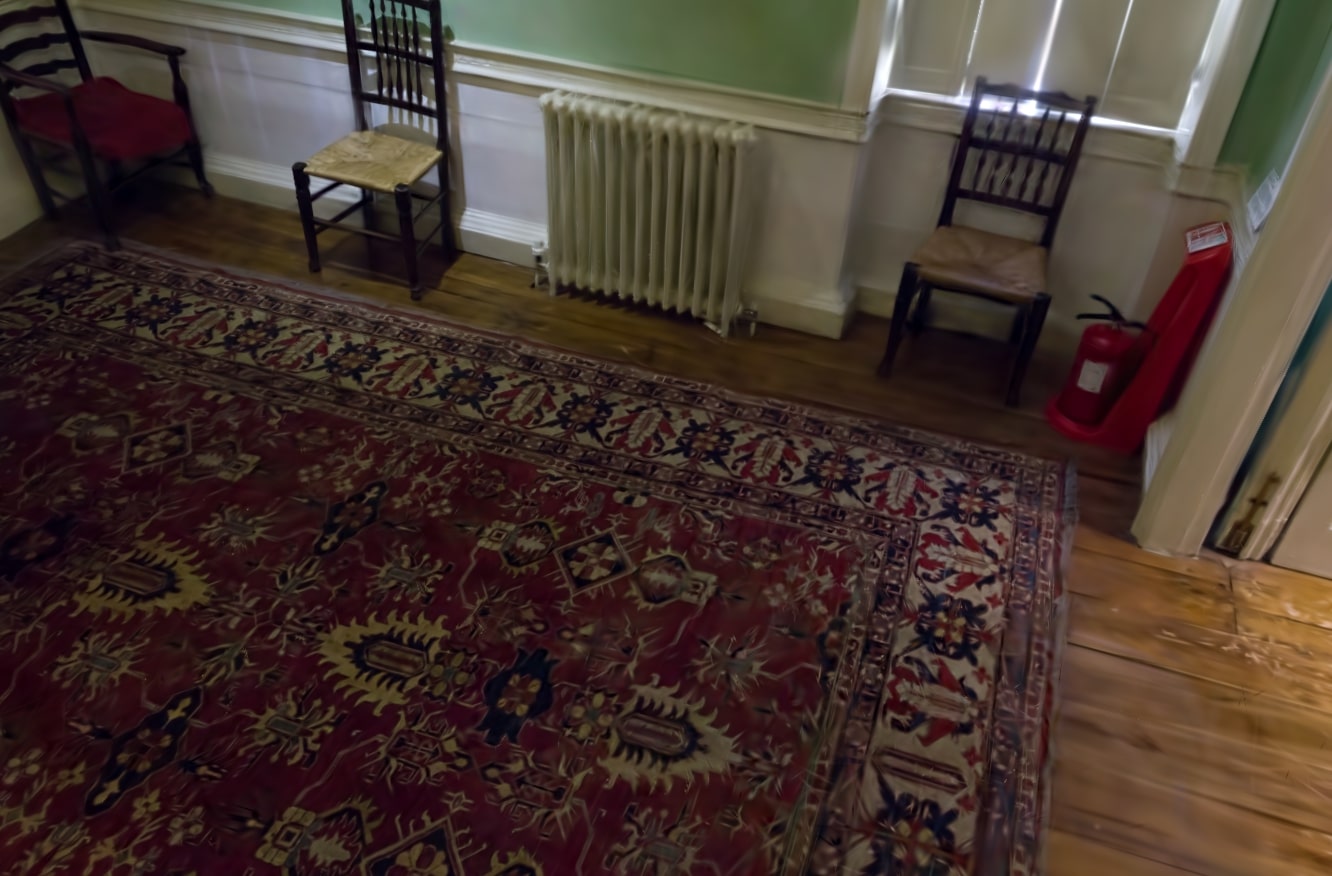}{2.5, 0.3}{4, 0.7} &
    \spyimage{red}{\resultfigwidth}{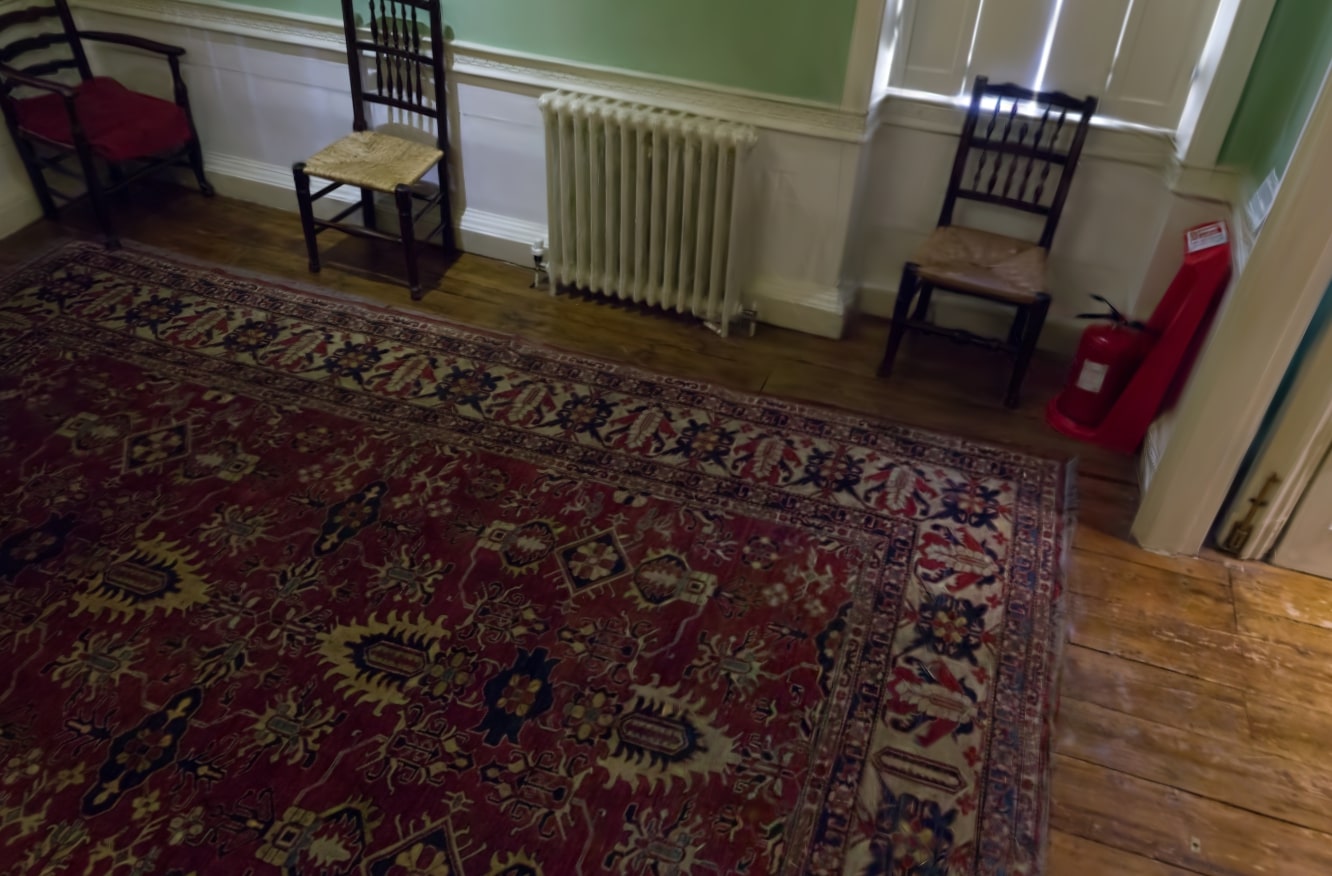}{2.5, 0.3}{4, 0.7} \\

    \raisebox{0.15\height}{\rotatebox{90}{\small Tanks\&Temples}} &
    \spyimage{red}{\resultfigwidth}{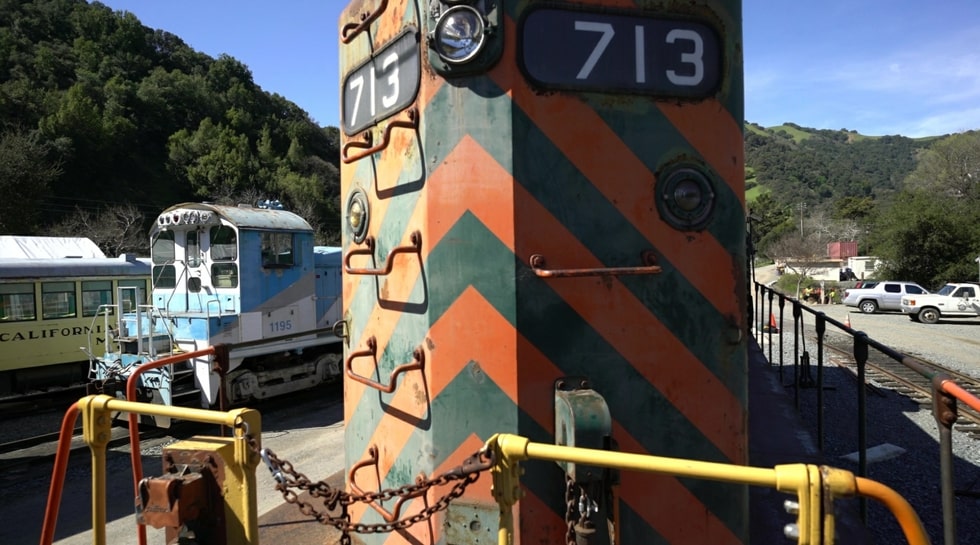}{2.1, 0.1}{1.1, 0.7} &
    \spyimage{red}{\resultfigwidth}{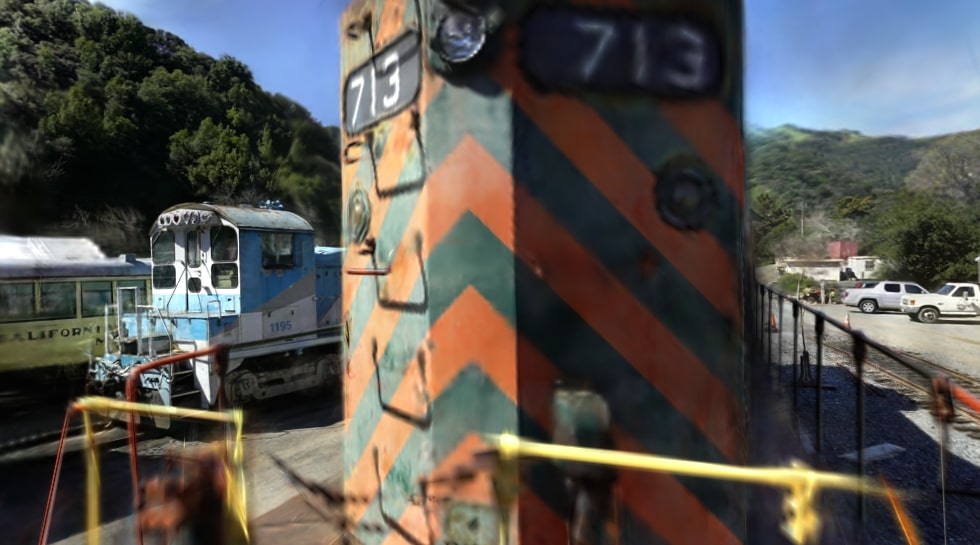}{2.1, 0.1}{1.1, 0.7} &
    \spyimage{red}{\resultfigwidth}{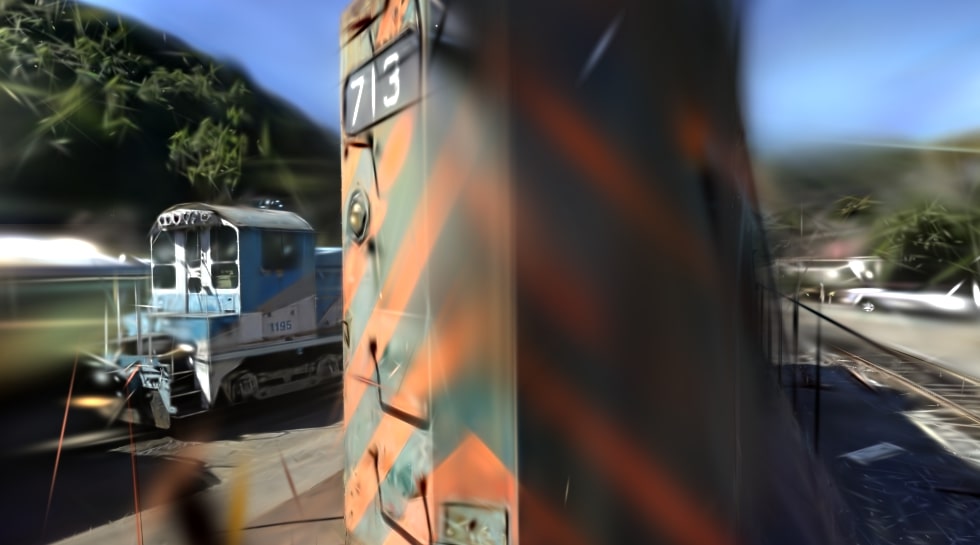}{2.1, 0.1}{1.1, 0.7} &
    \spyimage{red}{\resultfigwidth}{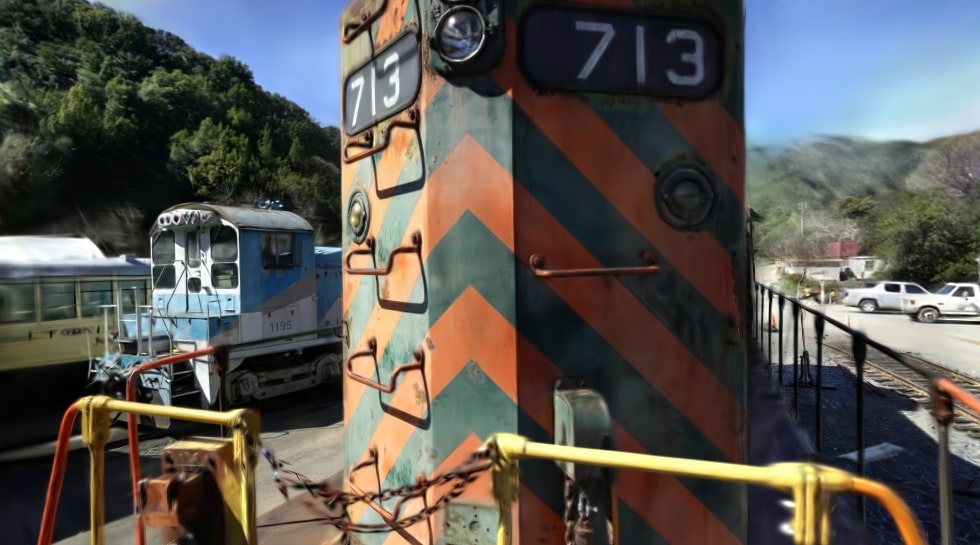}{2.1, 0.1}{1.1, 0.7} \\
\end{tabular}

\caption{
\label{fig:same-param-eval}
Adjusting BBSplat and GSTex to use the same budget as our method, results in significant visual degradation.}
\end{figure*}

\textbf{Qualitative Evaluation.}
In Figure~\ref{fig:default-eval},
we show visual results for the different models. We show
test views (i.e., not used for training) from one scene of each dataset.
Our method succeeds in reconstructing high-frequency details with high fidelity,
even with fewer parameters.
Similarly, Figure~\ref{fig:same-param-eval} displays visual results from the second experiment,
which highlight that other texturing methods struggle when constrained to the same parameter budget as ours.
GSTex uses a point cloud that is not trained with textured primitives in mind
and a static heuristic distribution of texels, which causes it to
not reconstruct well large parts of the scene.
BBSplat succeeds in geometrically reconstructing the scene, but exhibits texture stretching, which appears as blurriness.
This is due to the choice of mapping between intersection points and UV coordinates. 

\textbf{Quantitative Evaluation.}
For our quantitative comparisons,
we present results for three error metrics, SSIM, PSNR, and LPIPS \cite{lpips},
that are typically used to evaluate NVS.
Additionally,
we report the number of primitives and
texels used,
which provides a precise measure of the resources required for each method
and demonstrates whether and how much textures contribute to the scene reconstruction.
\revision{rev9584 scope of the paper, primitive vs parameters}{
Finally,
since the models' capacity is directly tied to the number of parameters used,
we include the relevant column so that we can make fair comparisons and draw meaningful conclusions.
}

\revision{rev4160}{
The formula for the number of parameters differs slightly among the models.
}
For 2DGS, GSTex each primitive has 58 parameters (3 for position, 2 for scale, 4 for rotation, 1 for opacity, 48 for color),
while for BBSplat this number is 57,
as it lacks opacity.
In our method,
in addition to the parameters of 2DGS,
we include one additional per primitive for the texel size,
pushing the number to 59 parameters.
Regarding the parameters coming from texels,
GSTex and our method have 3 per texel,
while BBSplat has one additional,
corresponding to the alpha channel.

\begin{table*}[!ht]
\footnotesize
\setlength\tabcolsep{2pt}
\centering
\begin{tabular}{@{}>{\raggedright\arraybackslash}p{1.3cm}|>{\centering\arraybackslash}p{0.8cm}>{\centering\arraybackslash}p{0.8cm}>{\centering\arraybackslash}p{0.8cm}>{\raggedleft\arraybackslash}p{0.7cm}>{\centering\arraybackslash}p{0.9cm}>{\centering\arraybackslash}p{0.6cm}|>{\centering\arraybackslash}p{0.8cm}>{\centering\arraybackslash}p{0.8cm}>{\centering\arraybackslash}p{0.8cm}>{\raggedleft\arraybackslash}p{0.7cm}>{\centering\arraybackslash}p{0.9cm}>{\centering\arraybackslash}p{0.6cm}|>{\centering\arraybackslash}p{0.8cm}>{\centering\arraybackslash}p{0.8cm}>{\centering\arraybackslash}p{0.8cm}>{\raggedleft\arraybackslash}p{0.7cm}>{\centering\arraybackslash}p{0.9cm}>{\centering\arraybackslash}p{0.6cm}}
 & \multicolumn{6}{c|}{DeepBlending} & \multicolumn{6}{c|}{Mip-Nerf-360} & \multicolumn{6}{c}{Tanks\&Temples} \\
 & SSIM$\uparrow$ & PSNR$\uparrow$ & LPIPS$\downarrow$ & Texels & Params & FPS & SSIM$\uparrow$ & PSNR$\uparrow$ & LPIPS$\downarrow$ & Texels & Params & FPS & SSIM$\uparrow$ & PSNR$\uparrow$ & LPIPS$\downarrow$ & Texels & Params & FPS \\
\midrule
Points40k & 0.890 & 28.78 & 0.346 & 13.2M & 41.9M & 121 & 0.758 & 25.80 & 0.304 & 42.2M & 128.9M & 97 & 0.809 & 22.74 & 0.246 & 15.4M & 48.4M & 162 \\
Points80k & {\cellcolor{tabthird}} 0.898 & {\cellcolor{tabthird}} 29.56 & {\cellcolor{tabthird}} 0.328 & 15.5M & 51.2M & 88 & {\cellcolor{tabthird}} 0.777 & {\cellcolor{tabthird}} 26.39 & {\cellcolor{tabthird}} 0.285 & 43.9M & 136.5M & 79 & {\cellcolor{tabthird}} 0.823 & {\cellcolor{tabthird}} 23.25 & {\cellcolor{tabthird}} 0.235 & 12.3M & 41.7M & 108 \\
Points120k & {\cellcolor{tabsecond}} 0.903 & {\cellcolor{tabsecond}} 29.88 & {\cellcolor{tabsecond}} 0.315 & 17.9M & 60.7M & 75 & {\cellcolor{tabsecond}} 0.785 & {\cellcolor{tabsecond}} 26.68 & {\cellcolor{tabsecond}} 0.275 & 45.0M & 142.0M & 68 & {\cellcolor{tabsecond}} 0.830 & {\cellcolor{tabsecond}} 23.34 & {\cellcolor{tabsecond}} 0.232 & 10.9M & 39.6M & 96 \\
Points160k & {\cellcolor{tabfirst}} 0.905 & {\cellcolor{tabfirst}} 29.98 & {\cellcolor{tabfirst}} 0.308 & 19.9M & 69.1M & 56 & {\cellcolor{tabfirst}} 0.791 & {\cellcolor{tabfirst}} 26.88 & {\cellcolor{tabfirst}} 0.269 & 45.7M & 146.6M & 46 & {\cellcolor{tabfirst}} 0.833 & {\cellcolor{tabfirst}} 23.44 & {\cellcolor{tabfirst}} 0.227 & 10.5M & 40.8M & 88 \\
\end{tabular}

\caption{\label{tab:point-budget}
We run our technique with a limited primitive budget.
An increased primitive count increases quality,
without proportionally increasing the number of texels and parameters.
}
\end{table*}

In Table~\ref{tab:default-eval},
we compare methods trained with default settings.
In most cases,
our approach achieves competitive or superior quality across all three metrics,
while requiring a lower number of trainable parameters. 
Focusing on the composition of these parameters,
our converged models have significantly fewer,
but more expressive primitives than the baseline as illustrated in Fig.~\ref{fig:teaser} (left).
This highlights the ability of our model to account for the difference in complexity between geometry and appearance.
In contrast, GSTex uses a pretrained 2DGS point cloud and has a fixed budget of texels, distributed at initialization time; this results in limited usage of texture. 
This is in part because it starts with a converged set of primitives from 2DGS that is not well adapted to a solution with texture.
On the contrary, we optimize both primitives and textures from scratch.

Moreover,
while our primitive count is close to BBSplat's,
the total number of parameters used is significantly lower than theirs.
This can be attributed to our content-aware texel size determination,
which dynamically allocates model capacity to regions according to their needs,
as opposed to the fixed texel to primitive ratio that BBSplat imposes.

To emphasize the importance of our content-aware texturing approach,
we next compare the other texturing methods by fixing both the primitive and texel budgets to ours.
In GSTex this is feasible,
because it allows the distribution of a given texel budget over a pretrained 2DGS point cloud.
We first trained 2DGS with the specified primitive budget and then gave the texel budget as input to GSTex.
However,
in BBSplat, controlling both the primitive and texel budgets is not possible,
as the method imposes a fixed ratio between the two,
determined by the texture resolution (16x16).
Therefore,
we calculated the number of primitives that would result in the same number of trainable parameters.
Table~\ref{tab:same-param-eval},
reports these results under these fixed parameters.
Note that BBSplat uses a skybox to model background and distant objects in outdoor scenes,
which was not taken into consideration when calculating the number of primitives.
This accounts for the slight discrepancy in the number of parameters.
Both the other methods observe a noticeable to significant drop in performance.
This is possibly due to the fact that model capacity is not distributed efficiently both across primitives and across parts of the scene.
In this setup, our method performs consistently better on average than previous solutions.

As a third experiment,
we run our method with a low, fixed primitive budget,
by skipping the splitting procedure whenever the model exceeds it.
The adaptive texel size strategies were left unchanged.
Table~\ref{tab:point-budget} shows the results.
As expected,
the performance gets better with an increasing primitive count,
as the use of primitives with a fixed gaussian falloff prevents faithfully reconstructing sharp edges with sparser point clouds.
However,
we note that the number of texels does not grow proportionally to the number of primitives
or even diminishes, in the case of Tanks~\&~Temples.
This is a direct result of our choice of texture coordinate mapping function and the adaptive texel size determination strategy.
Our separation of geometric and appearance parameters allows our models to automatically achieve a balance between the two that is appropriate for the scene.
\TODO{add BBSplat to show the proportional rise of parameters. Add the matplot graph}




\section{Limitations and Discussion}

Our method presents a good balance of resources used for textured Gaussian Splatting, allowing a smooth variation between number of primitives and number of texels used to represent a scene. However, like all other methods that build on 2DGS, we do not achieve the quality of 3DGS. Using 3D primitives with texture raises interesting questions about how to represent texture: should one use a 2D texture on a plane, or rather a voxel or hash-grid in 3D ? While Textured-GS~\cite{textured3dgs} proposes a first solution using the former approach, the analysis presented in the paper is insufficient to determine how well the choices made perform compared to our solution.

We do not have any special treatment for anti-aliasing. For NVS, the user can only navigate freely within -- or at least close -- to the convex hull of the input cameras. Since we choose a $t_2p_r$ value that is at least 2, aliasing will rarely by occuring in this range of viewpoints. Nonetheless, a complete solution for anti-aliasing would be an interesting avenue of future work.

We did not investigate the use of dedicated GPU hardware capabilities for texture. Given that our implementation of textured Gaussian Splatting uses a custom CUDA renderer, it is unclear how beneficial this would actually be. However, for the case, e.g., of WebGL renderers~\cite{webglviewer}, this may be a much more interesting direction that could allow accelerated rendering, especially in the case of low-end hardware. 

\section{Conclusion}

We have presented a new representation for textured 2D Gaussian Splatting, that is driven by scene content. By adaptively choosing texel size to fit the content of the scene, and carefully balancing resources used with our resolution-aware spltting approach, we provide a versatile method that allows users to choose between more primitives or more texture while preserving image quality. Our method provides an additional point in the design space of primitive-based NVS algorithms, building on the long tradition of texture mapping in CG rendering.

\section{Acknowledgements}
This work was funded by the European Research Council (ERC) Advanced Grant NERPHYS, number 101141721 \url{https://project.inria.fr/nerphys}.
The authors are grateful to the OPAL infrastructure of the Université Côte d'Azur for providing resources and support, as well as Adobe and NVIDIA for software and hardware donations.
F. Durand acknowledges funding from Google,  Amazon, and MIT-GIST.
The authors thank the anonymous reviewers for their valuable feedback.
\printbibliography

@String{tog = "ACM TOG"}

@ARTICLE{3dgs,title={3D Gaussian Splatting for Real-Time Radiance Field Rendering},year={2023},author={Bernhard Kerbl and Georgios Kopanas and Thomas Leimkühler and George Drettakis},doi={10.1145/3592433},pmid={null},pmcid={null},mag_id={4385318467},journal={ACM Transactions on Graphics}}

@article{reduced3dgs,
author = {Papantonakis, Panagiotis and Kopanas, Georgios and Kerbl, Bernhard and Lanvin, Alexandre and Drettakis, George},
title = {Reducing the Memory Footprint of 3D Gaussian Splatting},
year = {2024},
issue_date = {May 2024},
publisher = {Association for Computing Machinery},
address = {New York, NY, USA},
volume = {7},
number = {1},
url = {https://doi.org/10.1145/3651282},
doi = {10.1145/3651282},
journal = {Proc. ACM Comput. Graph. Interact. Tech.},
month = may,
articleno = {16},
numpages = {17}
}

@article{nerf,
author = {Mildenhall, Ben and Srinivasan, Pratul P. and Tancik, Matthew and Barron, Jonathan T. and Ramamoorthi, Ravi and Ng, Ren},
title = {NeRF: representing scenes as neural radiance fields for view synthesis},
year = {2021},
issue_date = {January 2022},
publisher = {Association for Computing Machinery},
address = {New York, NY, USA},
volume = {65},
number = {1},
issn = {0001-0782},
url = {https://doi.org/10.1145/3503250},
doi = {10.1145/3503250},
journal = {Commun. ACM},
month = dec,
pages = {99–106},
numpages = {8}
}

@InProceedings{mipnerf,
    author    = {Barron, Jonathan T. and Mildenhall, Ben and Tancik, Matthew and Hedman, Peter and Martin-Brualla, Ricardo and Srinivasan, Pratul P.},
    title     = {Mip-NeRF: A Multiscale Representation for Anti-Aliasing Neural Radiance Fields},
    booktitle = {Proceedings of the IEEE/CVF International Conference on Computer Vision (ICCV)},
    month     = {October},
    year      = {2021},
    pages     = {5855-5864}
}

@InProceedings{mipnerf360,
    author    = {Barron, Jonathan T. and Mildenhall, Ben and Verbin, Dor and Srinivasan, Pratul P. and Hedman, Peter},
    title     = {Mip-NeRF 360: Unbounded Anti-Aliased Neural Radiance Fields},
    booktitle = {Proceedings of the IEEE/CVF Conference on Computer Vision and Pattern Recognition (CVPR)},
    month     = {June},
    year      = {2022},
    pages     = {5470-5479}
}

@article{instant-ngp,
author = {M\"{u}ller, Thomas and Evans, Alex and Schied, Christoph and Keller, Alexander},
title = {Instant neural graphics primitives with a multiresolution hash encoding},
year = {2022},
issue_date = {July 2022},
publisher = {Association for Computing Machinery},
address = {New York, NY, USA},
volume = {41},
number = {4},
issn = {0730-0301},
url = {https://doi.org/10.1145/3528223.3530127},
doi = {10.1145/3528223.3530127},
journal = {ACM Trans. Graph.},
month = jul,
articleno = {102},
numpages = {15}
}

@inproceedings{tensorf,
author = {Chen, Anpei and Xu, Zexiang and Geiger, Andreas and Yu, Jingyi and Su, Hao},
title = {TensoRF: Tensorial Radiance Fields},
year = {2022},
isbn = {978-3-031-19823-6},
publisher = {Springer-Verlag},
address = {Berlin, Heidelberg},
url = {https://doi.org/10.1007/978-3-031-19824-3_20},
doi = {10.1007/978-3-031-19824-3_20},
booktitle = {Computer Vision – ECCV 2022: 17th European Conference, Tel Aviv, Israel, October 23–27, 2022, Proceedings, Part XXXII},
pages = {333–350},
numpages = {18},
location = {Tel Aviv, Israel}
}

@InProceedings{zipnerf,
    author    = {Barron, Jonathan T. and Mildenhall, Ben and Verbin, Dor and Srinivasan, Pratul P. and Hedman, Peter},
    title     = {Zip-NeRF: Anti-Aliased Grid-Based Neural Radiance Fields},
    booktitle = {Proceedings of the IEEE/CVF International Conference on Computer Vision (ICCV)},
    month     = {October},
    year      = {2023},
    pages     = {19697-19705}
}

@InProceedings{pixelnerf,
    author    = {Yu, Alex and Ye, Vickie and Tancik, Matthew and Kanazawa, Angjoo},
    title     = {pixelNeRF: Neural Radiance Fields From One or Few Images},
    booktitle = {Proceedings of the IEEE/CVF Conference on Computer Vision and Pattern Recognition (CVPR)},
    month     = {June},
    year      = {2021},
    pages     = {4578-4587}
}

@InProceedings{refnerf,
    author    = {Verbin, Dor and Hedman, Peter and Mildenhall, Ben and Zickler, Todd and Barron, Jonathan T. and Srinivasan, Pratul P.},
    title     = {Ref-NeRF: Structured View-Dependent Appearance for Neural Radiance Fields},
    booktitle = {Proceedings of the IEEE/CVF Conference on Computer Vision and Pattern Recognition (CVPR)},
    month     = {June},
    year      = {2022},
    pages     = {5491-5500}
}

@InProceedings{regnerf,
    author    = {Niemeyer, Michael and Barron, Jonathan T. and Mildenhall, Ben and Sajjadi, Mehdi S. M. and Geiger, Andreas and Radwan, Noha},
    title     = {RegNeRF: Regularizing Neural Radiance Fields for View Synthesis From Sparse Inputs},
    booktitle = {Proceedings of the IEEE/CVF Conference on Computer Vision and Pattern Recognition (CVPR)},
    month     = {June},
    year      = {2022},
    pages     = {5480-5490}
}

@INPROCEEDINGS{neuralangelo,
  author={Li, Zhaoshuo and Müller, Thomas and Evans, Alex and Taylor, Russell H. and Unberath, Mathias and Liu, Ming-Yu and Lin, Chen-Hsuan},
  booktitle={2023 IEEE/CVF Conference on Computer Vision and Pattern Recognition (CVPR)}, 
  title={Neuralangelo: High-Fidelity Neural Surface Reconstruction}, 
  year={2023},
  volume={},
  number={},
  pages={8456-8465},
  doi={10.1109/CVPR52729.2023.00817}}

@article{adaptiveshells,
author = {Wang, Zian and Shen, Tianchang and Nimier-David, Merlin and Sharp, Nicholas and Gao, Jun and Keller, Alexander and Fidler, Sanja and M\"{u}ller, Thomas and Gojcic, Zan},
title = {Adaptive Shells for Efficient Neural Radiance Field Rendering},
year = {2023},
issue_date = {December 2023},
publisher = {Association for Computing Machinery},
address = {New York, NY, USA},
volume = {42},
number = {6},
issn = {0730-0301},
url = {https://doi.org/10.1145/3618390},
doi = {10.1145/3618390},
journal = {ACM Trans. Graph.},
month = dec,
articleno = {260},
numpages = {15}
}

@inproceedings{bakedsdf,
author = {Yariv, Lior and Hedman, Peter and Reiser, Christian and Verbin, Dor and Srinivasan, Pratul P. and Szeliski, Richard and Barron, Jonathan T. and Mildenhall, Ben},
title = {BakedSDF: Meshing Neural SDFs for Real-Time View Synthesis},
year = {2023},
isbn = {9798400701597},
publisher = {Association for Computing Machinery},
address = {New York, NY, USA},
url = {https://doi.org/10.1145/3588432.3591536},
doi = {10.1145/3588432.3591536},
booktitle = {ACM SIGGRAPH 2023 Conference Proceedings},
articleno = {46},
numpages = {9},
location = {Los Angeles, CA, USA},
series = {SIGGRAPH '23}
}

@InProceedings{mipsplatting,
    author    = {Yu, Zehao and Chen, Anpei and Huang, Binbin and Sattler, Torsten and Geiger, Andreas},
    title     = {Mip-Splatting: Alias-free 3D Gaussian Splatting},
    booktitle = {Proceedings of the IEEE/CVF Conference on Computer Vision and Pattern Recognition (CVPR)},
    month     = {June},
    year      = {2024},
    pages     = {19447-19456}
}

@article{malarz2025gaussian,
  title={Gaussian splatting with NeRF-based color and opacity},
  author={Malarz, Dawid and Smolak-Dy{\.z}ewska, Weronika and Tabor, Jacek and Tadeja, S{\l}awomir and Spurek, Przemys{\l}aw},
  journal={Computer Vision and Image Understanding},
  volume={251},
  pages={104273},
  year={2025},
  publisher={Elsevier}
}

@inproceedings{specgaussian,
 author = {Yang, Ziyi and Gao, Xinyu and Sun, Yang-Tian and Huang, Yi-Hua and Lyu, Xiaoyang and Zhou, Wen and Jiao, Shaohui and Qi, Xiaojuan and Jin, Xiaogang},
 booktitle = {Advances in Neural Information Processing Systems},
 editor = {A. Globerson and L. Mackey and D. Belgrave and A. Fan and U. Paquet and J. Tomczak and C. Zhang},
 pages = {61192--61216},
 publisher = {Curran Associates, Inc.},
 title = {Spec-Gaussian: Anisotropic View-Dependent Appearance for 3D Gaussian Splatting},
 url = {https://proceedings.neurips.cc/paper_files/paper/2024/file/708e0d691a22212e1e373dc8779cbe53-Paper-Conference.pdf},
 volume = {37},
 year = {2024}
}

@article{yuksel2010mesh,
  title={Mesh colors},
  author={Yuksel, Cem and Keyser, John and House, Donald H},
  journal={ACM Transactions on Graphics (TOG)},
  volume={29},
  number={2},
  pages={1--11},
  year={2010},
  publisher={ACM New York, NY, USA}
}

@article{mildenhall2019local,
  title={Local light field fusion: Practical view synthesis with prescriptive sampling guidelines},
  author={Mildenhall, Ben and Srinivasan, Pratul P and Ortiz-Cayon, Rodrigo and Kalantari, Nima Khademi and Ramamoorthi, Ravi and Ng, Ren and Kar, Abhishek},
  journal={ACM Transactions on Graphics (ToG)},
  volume={38},
  number={4},
  pages={1--14},
  year={2019},
  publisher={ACM New York, NY, USA}
}

@article{zhou2018stereo,
author = {Zhou, Tinghui and Tucker, Richard and Flynn, John and Fyffe, Graham and Snavely, Noah},
title = {Stereo magnification: learning view synthesis using multiplane images},
year = {2018},
issue_date = {August 2018},
publisher = {Association for Computing Machinery},
address = {New York, NY, USA},
volume = {37},
number = {4},
issn = {0730-0301},
url = {https://doi.org/10.1145/3197517.3201323},
doi = {10.1145/3197517.3201323},
journal = {ACM Trans. Graph.},
month = jul,
articleno = {65},
numpages = {12}
}

@article{li2018optcuts,
  title={Optcuts: Joint optimization of surface cuts and parameterization},
  author={Li, Minchen and Kaufman, Danny M and Kim, Vladimir G and Solomon, Justin and Sheffer, Alla},
  journal={ACM transactions on graphics (TOG)},
  volume={37},
  number={6},
  pages={1--13},
  year={2018},
  publisher={ACM New York, NY, USA}
}

@article{hormann2007mesh,
  title={Mesh parameterization: Theory and practice},
  author={Hormann, Kai and L{\'e}vy, Bruno and Sheffer, Alla},
  year={2007}
}

@book{catmull1974subdivision,
  title={A subdivision algorithm for computer display of curved surfaces},
  author={Catmull, Edwin Earl},
  year={1974},
  publisher={The University of Utah}
}

@inproceedings{tewari2022advances,
  title={Advances in neural rendering},
  author={Tewari, Ayush and Thies, Justus and Mildenhall, Ben and Srinivasan, Pratul and Tretschk, Edgar and Yifan, Wang and Lassner, Christoph and Sitzmann, Vincent and Martin-Brualla, Ricardo and Lombardi, Stephen and others},
  booktitle={Computer Graphics Forum},
  volume={41},
  number={2},
  pages={703--735},
  year={2022},
  organization={Wiley Online Library}
}

@article{zhang2020nerf++,
  title={Nerf++: Analyzing and improving neural radiance fields},
  author={Zhang, Kai and Riegler, Gernot and Snavely, Noah and Koltun, Vladlen},
  journal={arXiv preprint arXiv:2010.07492},
  year={2020}
}

@inproceedings{sharma2024volumetric,
  title={Volumetric rendering with baked quadrature fields},
  author={Sharma, Gopal and Rebain, Daniel and Yi, Kwang Moo and Tagliasacchi, Andrea},
  booktitle={European Conference on Computer Vision},
  pages={275--292},
  year={2024},
  organization={Springer}
}

@inproceedings{yuksel2019rethinking,
  title={Rethinking texture mapping},
  author={Yuksel, Cem and Lefebvre, Sylvain and Tarini, Marco},
  booktitle={Computer graphics forum},
  volume={38},
  number={2},
  pages={535--551},
  year={2019},
  organization={Wiley Online Library}
}

@article{mallett2020patch,
  title={Patch textures: Hardware support for mesh colors},
  author={Mallett, Ian and Seiler, Larry and Yuksel, Cem},
  journal={IEEE Transactions on Visualization and Computer Graphics},
  volume={28},
  number={7},
  pages={2710--2721},
  year={2020},
  publisher={IEEE}
}

@inproceedings{srinivasan2024nuvo,
  title={Nuvo: Neural uv mapping for unruly 3d representations},
  author={Srinivasan, Pratul P and Garbin, Stephan J and Verbin, Dor and Barron, Jonathan T and Mildenhall, Ben},
  booktitle={European Conference on Computer Vision},
  pages={18--34},
  year={2024},
  organization={Springer}
}

@inproceedings{uv3dgs,
author = {Xu, Tian-Xing and Hu, Wenbo and Lai, Yu-Kun and Shan, Ying and Zhang, Song-Hai},
title = {Texture-GS: Disentangling the Geometry and Texture for 3D Gaussian Splatting Editing},
year = {2024},
isbn = {978-3-031-72697-2},
publisher = {Springer-Verlag},
address = {Berlin, Heidelberg},
url = {https://doi.org/10.1007/978-3-031-72698-9_3},
doi = {10.1007/978-3-031-72698-9_3},
booktitle = {Computer Vision – ECCV 2024: 18th European Conference, Milan, Italy, September 29–October 4, 2024, Proceedings, Part XXV},
pages = {37–53},
numpages = {17},
location = {Milan, Italy}
}

@inproceedings{2dgs,
author = {Huang, Binbin and Yu, Zehao and Chen, Anpei and Geiger, Andreas and Gao, Shenghua},
title = {2D Gaussian Splatting for Geometrically Accurate Radiance Fields},
year = {2024},
isbn = {9798400705250},
publisher = {Association for Computing Machinery},
address = {New York, NY, USA},
url = {https://doi.org/10.1145/3641519.3657428},
doi = {10.1145/3641519.3657428},
booktitle = {ACM SIGGRAPH 2024 Conference Papers},
articleno = {32},
numpages = {11},
location = {Denver, CO, USA},
series = {SIGGRAPH '24}
}

@inproceedings{revising,
author = {Rota Bul\`{o}, Samuel and Porzi, Lorenzo and Kontschieder, Peter},
title = {Revising Densification in Gaussian Splatting},
year = {2024},
isbn = {978-3-031-73035-1},
publisher = {Springer-Verlag},
address = {Berlin, Heidelberg},
url = {https://doi.org/10.1007/978-3-031-73036-8_20},
doi = {10.1007/978-3-031-73036-8_20},
booktitle = {Computer Vision – ECCV 2024: 18th European Conference, Milan, Italy, September 29–October 4, 2024, Proceedings, Part LXIII},
pages = {347–362},
numpages = {16},
location = {Milan, Italy}
}

@inproceedings{3dgs-mcmc,
 author = {Kheradmand, Shakiba and Rebain, Daniel and Sharma, Gopal and Sun, Weiwei and Tseng, Yang-Che and Isack, Hossam and Kar, Abhishek and Tagliasacchi, Andrea and Yi, Kwang Moo},
 booktitle = {Advances in Neural Information Processing Systems},
 editor = {A. Globerson and L. Mackey and D. Belgrave and A. Fan and U. Paquet and J. Tomczak and C. Zhang},
 pages = {80965--80986},
 publisher = {Curran Associates, Inc.},
 title = {3D Gaussian Splatting as Markov Chain Monte Carlo},
 url = {https://proceedings.neurips.cc/paper_files/paper/2024/file/93be245fce00a9bb2333c17ceae4b732-Paper-Conference.pdf},
 volume = {37},
 year = {2024}
}

@INPROCEEDINGS{kilonerf,
          author = {Christian Reiser and Songyou Peng and Yiyi Liao and Andreas Geiger},
          title = {KiloNeRF: Speeding up Neural Radiance Fields with Thousands of Tiny MLPs},
          booktitle = {International Conference on Computer Vision (ICCV)},
          year = {2021}
        }

@inproceedings{dvxgo,
  author    = {Cheng Sun and Min Sun and Hwann{-}Tzong Chen},
  title     = {Direct Voxel Grid Optimization: Super-fast Convergence for Radiance Fields Reconstruction},
  booktitle = {CVPR},
  year      = {2022},
}

@inproceedings{plenoxels,
      title={Plenoxels: Radiance Fields without Neural Networks}, 
      author={Sara Fridovich-Keil and Alex Yu and Matthew Tancik and Qinhong Chen and Benjamin Recht and Angjoo Kanazawa},
      year={2022},
      booktitle={CVPR},
}

@misc{bbsplat,
      title={BillBoard Splatting (BBSplat): Learnable Textured Primitives for Novel View Synthesis}, 
      author={David Svitov and Pietro Morerio and Lourdes Agapito and Alessio Del Bue},
      year={2025},
      eprint={2411.08508},
      archivePrefix={arXiv},
      primaryClass={cs.CV},
      url={https://arxiv.org/abs/2411.08508}, 
}

@inproceedings{textured3dgs,
            title={Textured Gaussians for Enhanced 3D Scene Appearance Modeling},
            author={Brian Chao and Hung-Yu Tseng and Lorenzo Porzi and Chen Gao and Tuotuo Li and Qinbo Li and Ayush Saraf and Jia-Bin Huang and Johannes Kopf and Gordon Wetzstein and Changil Kim},
            year={2025},
            booktitle={CVPR}
}

@misc{supergaussians,
      title={SuperGaussians: Enhancing Gaussian Splatting Using Primitives with Spatially Varying Colors}, 
      author={Xu, Rui and Chen, Wenyue and Wang, Jiepeng and Liu, Yuan and Wang, Peng and Gao, Lin and Xin, Shiqing and Komura, Taku and Li, Xin and Wang, Wenping},
      year={2024},
      archivePrefix={arXiv},
      primaryClass={cs.LG},
}

@InProceedings{gstex,
    author    = {Rong, Victor and Chen, Jingxiang and Bahmani, Sherwin and Kutulakos, Kiriakos and Lindell, David},
    title     = {GStex: Per-Primitive Texturing of 2D Gaussian Splatting for Decoupled Appearance and Geometry Modeling},
    booktitle = {Proceedings of the Winter Conference on Applications of Computer Vision (WACV)},
    month     = {February},
    year      = {2025},
    pages     = {3508-3518}
}

@article{tnt,
author = {Knapitsch, Arno and Park, Jaesik and Zhou, Qian-Yi and Koltun, Vladlen},
title = {Tanks and temples: benchmarking large-scale scene reconstruction},
year = {2017},
issue_date = {August 2017},
publisher = {Association for Computing Machinery},
address = {New York, NY, USA},
volume = {36},
number = {4},
issn = {0730-0301},
url = {https://doi.org/10.1145/3072959.3073599},
doi = {10.1145/3072959.3073599},
journal = {ACM Trans. Graph.},
month = jul,
articleno = {78},
numpages = {13}
}

@article{db,
author = {Hedman, Peter and Philip, Julien and Price, True and Frahm, Jan-Michael and Drettakis, George and Brostow, Gabriel},
title = {Deep blending for free-viewpoint image-based rendering},
year = {2018},
issue_date = {December 2018},
publisher = {Association for Computing Machinery},
address = {New York, NY, USA},
volume = {37},
number = {6},
issn = {0730-0301},
url = {https://doi.org/10.1145/3272127.3275084},
doi = {10.1145/3272127.3275084},
journal = {ACM Trans. Graph.},
month = dec,
articleno = {257},
numpages = {15}
}

@InProceedings{lpips,
author = {Zhang, Richard and Isola, Phillip and Efros, Alexei A. and Shechtman, Eli and Wang, Oliver},
title = {The Unreasonable Effectiveness of Deep Features as a Perceptual Metric},
booktitle = {Proceedings of the IEEE Conference on Computer Vision and Pattern Recognition (CVPR)},
month = {June},
year = {2018}
}

@article{3dgszip,
    title={3DGS.zip: A survey on 3D Gaussian Splatting Compression Methods}, 
    author={Milena T. Bagdasarian and Paul Knoll and Yi-Hsin Li and Florian Barthel and Anna Hilsmann and 
            Peter Eisert and Wieland Morgenstern},
    journal={arXiv preprint arXiv:2407.09510},
    year={2024}, 
}

@misc{webglviewer,
    title={Splat Viewer},
    url={https://github.com/antimatter15/splat?tab=readme-ov-file},
    author={Kevin Kwok}
}

\begin{table*}[!h]
\footnotesize
\setlength\tabcolsep{1.5pt}
\centering
\scalebox{0.95}{
\begin{tabular}{@{}>{\raggedright\arraybackslash}p{1.2cm}|>{\centering\arraybackslash}p{0.8cm}>{\centering\arraybackslash}p{0.8cm}>{\centering\arraybackslash}p{0.8cm}>{\raggedleft\arraybackslash}p{0.8cm}>{\raggedleft\arraybackslash}p{0.7cm}>{\centering\arraybackslash}p{0.9cm}>{\centering\arraybackslash}p{0.55cm}|>{\centering\arraybackslash}p{0.8cm}>{\centering\arraybackslash}p{0.8cm}>{\centering\arraybackslash}p{0.8cm}>{\raggedleft\arraybackslash}p{0.8cm}>{\raggedleft\arraybackslash}p{0.7cm}>{\centering\arraybackslash}p{0.9cm}>{\centering\arraybackslash}p{0.55cm}|>{\centering\arraybackslash}p{0.8cm}>{\centering\arraybackslash}p{0.8cm}>{\centering\arraybackslash}p{0.8cm}>{\raggedleft\arraybackslash}p{0.8cm}>{\raggedleft\arraybackslash}p{0.7cm}>{\centering\arraybackslash}p{0.9cm}>{\centering\arraybackslash}p{0.55cm}}
 & \multicolumn{7}{c|}{DeepBlending} & \multicolumn{7}{c|}{Mip-Nerf-360} & \multicolumn{7}{c}{Tanks\&Temples} \\
 & SSIM$\uparrow$ & PSNR$\uparrow$ & LPIPS$\downarrow$ & Points & Texels & Params & FPS & SSIM$\uparrow$ & PSNR$\uparrow$ & LPIPS$\downarrow$ & Points & Texels & Params & FPS & SSIM$\uparrow$ & PSNR$\uparrow$ & LPIPS$\downarrow$ & Points & Texels & Params & FPS \\
\midrule
Default & 0.907 & 30.03 & 0.303 & 222K & 21.6M & 78.1M & 70 & 0.795 & 27.00 & 0.263 & 218K & 46.6M & 152.6M & 67 & 0.835 & 23.43 & 0.225 & 164K & 10.4M & 41.1M & 121 \\
Higher $\tau_\text{ds}$ & 0.903 & 29.90 & 0.324 & 196K & 7.9M & 35.4M & 67 & 0.791 & 26.97 & 0.277 & 233K & 33.8M & 115.1M & 60 & 0.829 & 23.40 & 0.246 & 165K & 6.6M & 29.4M & 118 \\
Lower $\tau_{tr}$ & 0.907 & 30.08 & 0.300 & 316K & 22.0M & 84.8M & 66 & 0.812 & 27.30 & 0.244 & 542K & 46.3M & 170.8M & 52 & 0.842 & 23.67 & 0.217 & 281K & 10.9M & 49.3M & 104 \\
\end{tabular}
}
\caption{\label{tab:extra-models} \revision{}{We provide results for two extra models, trained with different $\tau_{ds}$ and $\tau_{tr}$ to demonstrate the effect of these hyperparameters in the final model.}}

\end{table*}

\begin{table*}[!h]
\footnotesize
\setlength\tabcolsep{1.5pt}
\centering
\begin{tabular}{@{}>{\raggedright\arraybackslash}p{1.3cm}|>{\centering\arraybackslash}p{0.8cm}>{\centering\arraybackslash}p{0.8cm}>{\centering\arraybackslash}p{0.8cm}>{\raggedleft\arraybackslash}p{0.9cm}>{\raggedleft\arraybackslash}p{0.8cm}>{\centering\arraybackslash}p{1.3cm}>{\centering\arraybackslash}p{0.5cm}>{\raggedleft\arraybackslash}p{1cm}>{\centering\arraybackslash}p{1.3cm}}
Scene & SSIM$\uparrow$ & PSNR$\uparrow$ & LPIPS$\downarrow$ & Points & Texels & Params & FPS & Mem & \multirow{2}{4em}{Texels/ Primitive} \\
& & & & & & & & & \\
\midrule
drjohnson & 0.906 & 29.78 & 0.314 & 293K & 20.8M & 79.7M & 77 & 318MB & 70.9 \\
playroom & 0.907 & 30.28 & 0.293 & 152K & 22.5M & 76.4M & 64 & 305MB & 147.7 \\
\midrule
bicycle & 0.726 & 24.22 & 0.254 & 186K & 83.9M & 262.7M & 84 & 1050MB & 450.5 \\
bonsai & 0.933 & 31.46 & 0.253 & 291K & 9.0M & 44.1M & 35 & 176MB & 30.8 \\
counter & 0.900 & 28.79 & 0.261 & 237K & 12.4M & 51.3M & 41 & 205MB & 52.2 \\
flowers & 0.531 & 20.22 & 0.392 & 177K & 61.3M & 194.5M & 74 & 777MB & 345.6 \\
garden & 0.832 & 26.56 & 0.156 & 164K & 51.7M & 164.7M & 112 & 658MB & 314.6 \\
kitchen & 0.915 & 30.90 & 0.173 & 288K & 9.6M & 45.9M & 40 & 183MB & 33.3 \\
room & 0.921 & 31.72 & 0.272 & 212K & 17.7M & 65.7M & 62 & 262MB & 83.3 \\
stump & 0.756 & 26.09 & 0.273 & 228K & 101.4M & 317.6M & 77 & 1270MB & 444.5 \\
treehill & 0.641 & 23.01 & 0.335 & 180K & 72.2M & 227.2M & 77 & 908MB & 398.8 \\
\midrule
train & 0.795 & 21.38 & 0.279 & 172K & 7.2M & 31.8M & 139 & 127MB & 41.7 \\
truck & 0.875 & 25.47 & 0.172 & 156K & 13.7M & 50.4M & 103 & 201MB & 87.3 \\
\end{tabular}
\caption{\label{tab:per-scene} \revision{}{Per-scene metrics of our model, grouped by their dataset.}}

\end{table*}
\appendix
\revision{}{
\section{Implementation Details}
In this section we provide some more details on the implementation.
}

\revision{}{
The texture grids are initialized with a value of $0$ on every channel
and start optimizing after 500 iterations.
To avoid running out of memory and to avoid early overfitting,
at 500 iterations,
we set $t_2p_r$ for each primivite so that their smallest axis is 8 texels wide,
in practice using the closest power of two value.
}
\revision{}{
The error calculation along with the
adaptive texel size determination and resolution-aware primitive management strategies run every 250 iterations,
until 25k iterations.
The threshold $\tau_{\mathrm{tr}}$ starts at $64$ and is progressively reduced to $32$ within 7000 iterations.
}
\revision{}{
We implement downscaling and upscaling using pytorch's interpolate function,
with a scale factor of 2,
which translates to reductions or increases of the number of texels by a factor of 4,
respectively.
Upscale is done using nearest neighbour interpolation,
so that we avoid changing the appearance of the texture,
that could disrupt that optimizer.
}

\revision{rev4160 6315}{
\section{Hyperparameter Tuning}
}
\revision{}{
We provide here some intuition on their impact of hyperparameters to the final model.
}

\revision{}{
The quantile of the error $E$ used in both texel size adaptation and primitive management routines affects how aggressively they act on the model.
A lower quantile means more aggressive changes as more primitives are potentially upscaled and/or split,
while a higher one has the opposite effect.
$\tau_{tr}$ controls how many primitives are split or not,
contributing to the total number of primitives.
Setting this hyperparameter to a low number leads to a model with more primitives,
as it is easier for primitives to fall above the texture resolution threshold.
Finally,
$\lambda_\text{texture}$ controls the variance of the texture maps,
with higher values leading to smoother results,
while downscale parameter $t_\text{ds}a$ affects how easily a texture map can get downscaled,
and thus having a simpler appearance.
Both of these hyperparameters have a direct effect on the number of texels.
A low $\lambda_\text{texture}$ and high  $t_\text{ds}$ result in models with few texels per primitive,
while the opposite configuration leads to more texels being used overall,
increasing the number of parameters used.
}
\begin{algorithm}[!htbp]
\SetAlgoLined
\SetAlgoNoLine
    \If{$iter \bmod 250 = 0$}{
       \ForAll{primitives}{
        Compute $E_i$\;
            \If{texture resolution > $\tau_{tr}$ and $E_i$ in top 90\%}{
                Split primitive along overflowing axes\;
            }
            \If{$\mathrm{t_2p_r} > 1$ and $E_i$ in top 90\%} {
                Decrease texel size (Upscale)\;
            }
            \ElseIf{$\mathcal{E}_d < \tau_\text{ds}$}{
                Increase texel size (Downscale)\;
            }
        }
    }
\caption{\revision{}{Texel Size Adaptation and Primitive Management Routine}}
\label{alg:routines}
\end{algorithm}

\revision{}{
Note that since the number of primitives and the number of texels are linked through our resolution-aware primitive management,
a change in one hyperparameter can have secondary, indirect effects.
For example,
encouraging the spawning of more primitives by using a lower $\tau_{tr}$ can lead to fewer texels,
since each primitive represents a more localized and therefore simpler part of the scene.
This is demonstrated in the third experiment  (Tab~\ref{tab:point-budget}) with Tanks\&Temples.
Our method is robust enough to operate with different ratios of primitive and texel budgets,
converging to good results and automatically adjusting the allocation of model capacity.
}
\revision{}{
In Tab.~\ref{tab:extra-models} we provide two additional models,
one with more primitives, as a result of a lower $\tau_{tr}$ and one with less texels,
as a result of a higher $\tau_\text{ds}$.
}

\end{document}